\documentclass{article}

\PassOptionsToPackage{numbers, sort&compress}{natbib}
\usepackage[preprint]{neurips_2026}

\usepackage[utf8]{inputenc} % allow utf-8 input
\usepackage[T1]{fontenc}    % use 8-bit T1 fonts
\usepackage{hyperref}       % hyperlinks

\usepackage{url}            % simple URL typesetting
\usepackage{booktabs}       % professional-quality tables
\usepackage{tabularx}
\usepackage{graphicx}
\usepackage{xcolor}
\usepackage{colortbl}
\usepackage{enumitem}
\usepackage{array}
\usepackage{ragged2e}
\usepackage{makecell}
\usepackage{caption}
\usepackage{adjustbox}
\usepackage{tcolorbox}
\usepackage{fontawesome5}
\usepackage{tikz}
\usepackage{soul}
\usepackage{setspace}
\usepackage{algorithm}   
\usepackage{algpseudocode} 
\usepackage{amsmath}
\usepackage[nameinlink,capitalize]{cleveref}
\tcbuselibrary{skins,breakable}
\usetikzlibrary{shadows}
\usepackage{amssymb}
\usepackage{placeins}
\usepackage{float}
\usepackage{xcolor}
\usepackage{tcolorbox}
\tcbuselibrary{breakable,skins}
\definecolor{pboxbg}{HTML}{F9FAFB}          % pale panel background
\definecolor{pboxframe}{HTML}{CBD5E1}       % soft blue-gray frame
\definecolor{ptitlebg}{HTML}{E8EDF4}        % muted title background
\definecolor{ptitlefg}{HTML}{1E3A5F}        % dark blue title text
\definecolor{pvarcolor}{HTML}{0D7377}       % dynamic variables
\definecolor{pphcolor}{HTML}{7C8DA5}        % placeholders
\definecolor{pcodecolor}{HTML}{374151}      % code text
\definecolor{pdesccolor}{HTML}{1F2937}      % description text

\newcommand{\pvar}[1]{{\ttfamily\color{pvarcolor}#1}}
\newcommand{\pph}[1]{{\sffamily\itshape\color{pphcolor}#1}}
\newcommand{\pcode}[1]{{\ttfamily\color{pcodecolor}#1}}
\newcommand{\promptgap}{\vspace{0.5em}}

\newtcolorbox{promptbox}[1][]{%
  enhanced,
  breakable,
  colback=pboxbg,
  colframe=pboxframe,
  coltitle=ptitlefg,
  colbacktitle=ptitlebg,
  fonttitle=\small\bfseries\sffamily,
  title={#1},
  toptitle=2mm,
  bottomtitle=2mm,
  arc=1.5mm,
  boxrule=0.4pt,
  titlerule=0pt,
  left=8pt,
  right=8pt,
  top=10pt,
  bottom=8pt,
  boxsep=2mm,
  drop shadow southeast={opacity=0.06,shadow xshift=0.6pt,shadow yshift=-0.6pt},
  before upper={\linespread{1.15}\selectfont\small\sffamily\color{pdesccolor}},
  before skip=12pt plus 3pt,
  after skip=12pt plus 3pt,
}

\newtcolorbox{choicebox}[1][]{%
  enhanced, breakable,
  colback=gray!4, colframe=gray!60!black,
  colbacktitle=gray!15, coltitle=black,
  fonttitle=\bfseries\sffamily,
  left=6pt, right=6pt, top=4pt, bottom=4pt,
  boxrule=0.6pt,
  attach boxed title to top left={xshift=10pt, yshift=-2mm},
  boxed title style={sharp corners, size=small},
  #1
}

\definecolor{caseStatic}{HTML}{E0F2FE}
\definecolor{caseDynamic}{HTML}{DCFCE7}
\definecolor{caseInterleaved}{HTML}{F3E8FF}
\definecolor{caseApplied}{HTML}{FFE7D6}
\definecolor{caseInk}{HTML}{253044}
\definecolor{caseMuted}{HTML}{667085}

\newtcolorbox{casecard}[2][]{%
  enhanced,
  colback=white,
  colframe=#2,
  boxrule=0.7pt,
  arc=2mm,
  left=5pt,
  right=5pt,
  top=5pt,
  bottom=5pt,
  before skip=5pt,
  after skip=5pt,
  #1
}

\newcommand{\casebadge}[2]{%
  \tikz[baseline=(text.base)]{
    \node[fill=#1, text=caseInk, rounded corners=5pt,
          inner xsep=6pt, inner ysep=2pt,
          font=\sffamily\bfseries\scriptsize] (text) {#2};
  }%
}

\newcommand{\skillmaptile}[8]{%
  \pgfmathsetmacro{\cx}{#1 + #3/2}%
  \pgfmathsetmacro{\imgy}{#2 - 0.10}%
  \pgfmathsetmacro{\titley}{#2 - #4 + 0.50}%
  \pgfmathsetmacro{\descy}{#2 - #4 + 0.28}%
  \node[anchor=north west,
        minimum width=#3cm,
        minimum height=#4cm,
        draw=cardline,
        fill=white,
        line width=0.16pt,
        rounded corners=3pt,
        inner sep=0pt,
        drop shadow={shadow xshift=0.060cm, shadow yshift=-0.060cm, opacity=0.16}] at (#1,#2) {};
  \node[anchor=north, inner sep=0pt] at (\cx,\imgy)
        {\includegraphics[width=\dimexpr#3cm-0.16cm\relax,height=#8cm,keepaspectratio,trim=0 24 0 24,clip]{#5}};
  \node[anchor=north, align=center, text=textmain,
        font=\sffamily\bfseries\fontsize{4.85pt}{5.15pt}\selectfont,
        text width=\dimexpr#3cm-0.36cm\relax]
        at (\cx,\titley) {#6};
  \node[anchor=north, align=center, text=textfaint,
        font=\sffamily\fontsize{4.15pt}{4.45pt}\selectfont,
        text width=\dimexpr#3cm-0.42cm\relax]
        at (\cx,\descy) {#7};
}

\newcommand{\skillmaptalltile}[8]{%
  \pgfmathsetmacro{\cx}{#1 + #3/2}%
  \pgfmathsetmacro{\imgy}{#2 - 0.10}%
  \pgfmathsetmacro{\titley}{#2 - #4 + 0.50}%
  \pgfmathsetmacro{\descy}{#2 - #4 + 0.28}%
  \node[anchor=north west,
        minimum width=#3cm,
        minimum height=#4cm,
        draw=cardline,
        fill=white,
        line width=0.16pt,
        rounded corners=3pt,
        inner sep=0pt,
        drop shadow={shadow xshift=0.060cm, shadow yshift=-0.060cm, opacity=0.16}] at (#1,#2) {};
  \node[anchor=north, inner sep=0pt] at (\cx,\imgy)
        {\includegraphics[width=\dimexpr#3cm-0.16cm\relax,height=#8cm,keepaspectratio,trim=0 24 0 24,clip]{#5}};
  \node[anchor=north, align=center, text=textmain,
        font=\sffamily\bfseries\fontsize{4.75pt}{5.05pt}\selectfont,
        text width=\dimexpr#3cm-0.45cm\relax]
        at (\cx,\titley) {#6};
  \node[anchor=north, align=center, text=textfaint,
        font=\sffamily\fontsize{4.08pt}{4.38pt}\selectfont,
        text width=\dimexpr#3cm-0.50cm\relax]
        at (\cx,\descy) {#7};
}

\definecolor{panelbg}{HTML}{F8FAFC}
\definecolor{panelstroke}{HTML}{DCE6ED}
\definecolor{cardline}{HTML}{DDE8EF}
\definecolor{accentblue}{HTML}{2F8EDB}
\definecolor{accentbluepale}{HTML}{EAF7FF}
\definecolor{softnavytitle}{HTML}{1D6078}
\colorlet{babyblue}{accentbluepale}
\colorlet{babyblueborder}{accentblue}
\colorlet{babybluedeep}{accentblue}

\definecolor{accentgreen}{HTML}{24B47E}
\definecolor{accentgreenpale}{HTML}{ECFBF4}
\definecolor{mintdark}{HTML}{147A52}
\colorlet{mintcream}{accentgreenpale}
\colorlet{mintborder}{accentgreen}
\colorlet{mintdeep}{accentgreen}

\definecolor{peachcream}{HTML}{FEE2E2}
\definecolor{peachborder}{HTML}{FCA5A5}
\definecolor{watermelonpink}{HTML}{F87171}
\definecolor{watermelondark}{HTML}{DC2626}

\definecolor{accentorange}{HTML}{F59E42}
\definecolor{accentorangepale}{HTML}{FFF3E8}
\definecolor{lavenderdark}{HTML}{A95D1D}
\colorlet{lavenderlight}{accentorangepale}
\colorlet{lavenderborder}{accentorange}
\colorlet{lavenderdeep}{accentorange}

\definecolor{buttercream}{HTML}{FEF9C3}
\definecolor{butterborder}{HTML}{FDE047}
\definecolor{butterdeep}{HTML}{EAB308}
\definecolor{butterdark}{HTML}{A16207}

\definecolor{cloudwhite}{HTML}{FEFEFE}
\definecolor{snowgray}{HTML}{F8FAFC}
\definecolor{mist}{HTML}{F1F5F9}
\definecolor{softborder}{HTML}{D3E2E8}
\definecolor{textsoft}{HTML}{445668}
\definecolor{textmain}{HTML}{243747}
\definecolor{textfaint}{HTML}{52677A}

\definecolor{rowpastel}{HTML}{F0F9FF}

\newcommand{\sectionlabel}[1]{%
  \tikz[baseline=(text.base)]{
    \node[fill=babyblue, text=softnavytitle, rounded corners=7pt,
          inner xsep=8pt, inner ysep=3.5pt,
          font=\sffamily\bfseries\footnotesize,
          draw=babyblueborder, line width=0.5pt] (text) {#1};
  }%
}

\usepackage{amsfonts}       % blackboard math symbols
\usepackage{nicefrac}       % compact symbols for 1/2, etc.
\usepackage{microtype}      % microtypography
\usepackage[table]{xcolor}
\usepackage{xcolor}         % colors
\usepackage{csquotes}
\usepackage{multirow}
\usepackage{graphicx}
\usepackage{xspace}
\title{Agent Skills Should Go Beyond Text: The Case for Visual Skills}

\newcommand{\NAME}{\textsc{Visual Skill}\xspace}
\newcommand{\SYSTEM}{\textsc{AutoVisualSkill}\xspace}
\definecolor{my_green}{RGB}{51,102,0}

\author{%
Binxiao Xu\textsuperscript{1}
,
Ruichuan An\textsuperscript{1}
,
Bocheng Zou\textsuperscript{2}
,
Hang Hua\textsuperscript{3,\ensuremath{\ddagger}}
\\
\textsuperscript{1}Peking University
,
\textsuperscript{2}University of Wisconsin
,
\textsuperscript{3}MIT-IBM Watson AI Lab
\\
\texttt{\{binxiao, ruichuan\}@pku.edu.cn}
,
\texttt{bochengz@cs.wisc.edu}
,
\texttt{hang.hua1@ibm.com}
}

\begin{document}

\maketitle

\begingroup

\renewcommand{\thefootnote}{\fnsymbol{footnote}}

\footnotetext[2]{Corresponding author.}

\endgroup

\begin{abstract}
Reusable skills are a key mechanism for extending agent capabilities, allowing agents to accumulate experience and solve increasingly complex tasks. 
Yet most existing skill-learning methods store reusable experience as text-only assets, such as instructions, reasoning traces, or summarized trajectories. 
We argue that this text-only paradigm creates a fundamental bottleneck for visual-centric tasks, where reusable knowledge often depends on spatial layout, visual grounding, fine-grained appearance, and localized state changes. 
To address this limitation, we propose \textbf{\NAME}, a multimodal skill paradigm that combines declarative textual logic with explicit visual support. 
We distinguish three reusable forms: static priors for stable spatial conventions, dynamic priors for in-situ visual working memory, and interleaved visual skills that bind ordered text steps to the source frames, screenshots, or page regions that justify them. 
Rather than only describing what to do, visual skills also encode where to look, how to inspect, and how to verify visual outcomes. 
To scale visual-skill construction, we introduce \textbf{\SYSTEM}, an automatic system that converts agent experience into reusable multimodal skills by preserving textual reasoning, spatial references, visual boundaries, and interaction patterns from task trajectories. 
Experiments on GUI and other visual-centric tasks show that visual skills consistently outperform text-only skills, particularly when success requires spatial correspondence, visual evidence, and state-aware interaction. 
These results support our central position: reusable agent skills should go beyond text and become multimodal assets for future multimodal agents. Resources available at \href{https://github.com/Little-Fridge/AutoVisualSkill}{\textcolor{my_green}{\fontfamily{cmtt}\selectfont{https://github.com/Little-Fridge/AutoVisualSkill}}}.
\end{abstract}

\begin{figure*}[t]
\centering
\includegraphics[width=\linewidth]{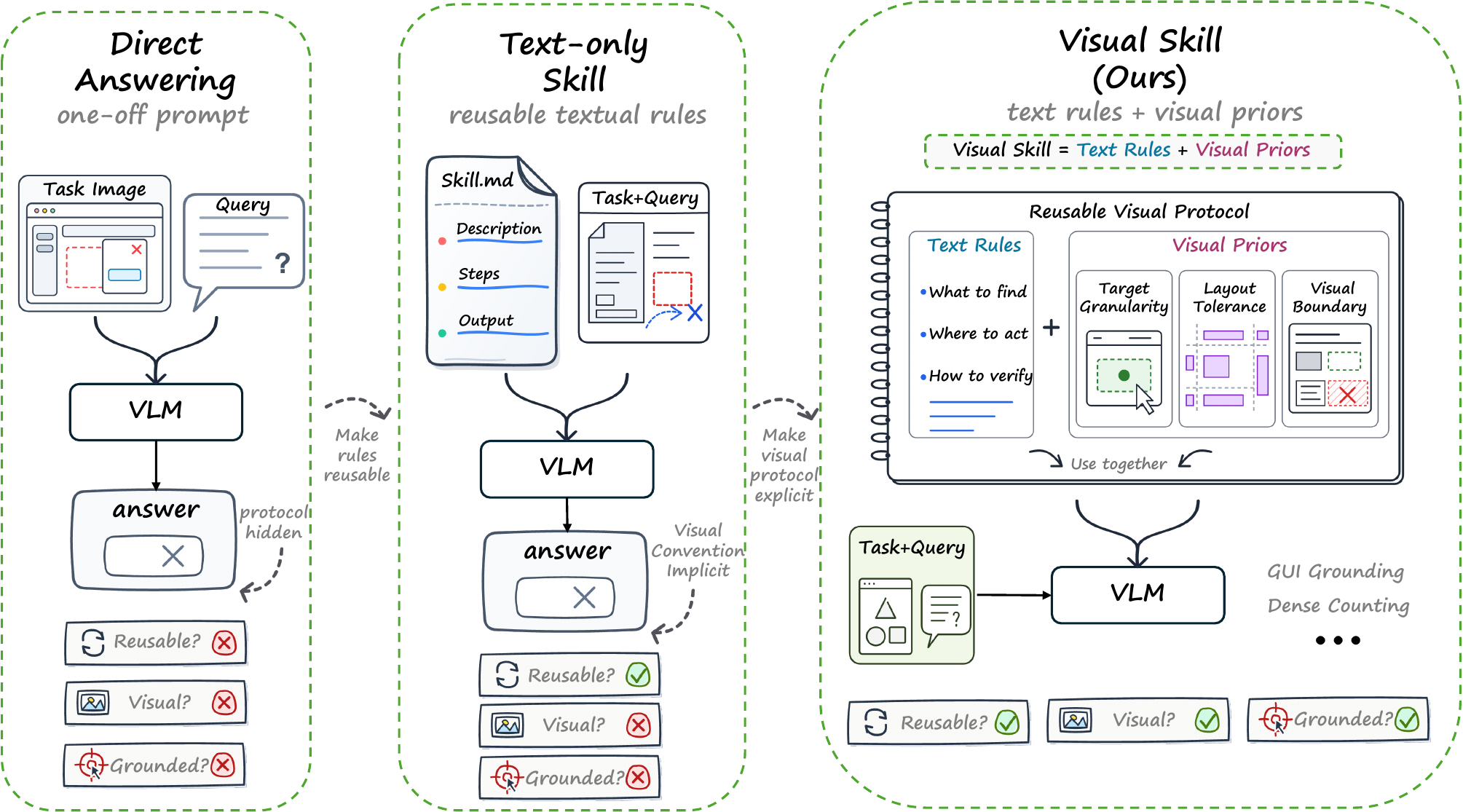}
\caption{\textbf{From text-only reuse to visual skills.}
Direct answering solves each visual task as a one-off prompt, while text-only skills can reuse rules but leave spatial conventions implicit.
\NAME combines reusable text rules with explicit visual priors, making the visual protocol visible, reusable, and grounded.}
\label{fig:teaser}
\vspace{-2mm}
\end{figure*}

\begin{figure*}[t]
\centering
\resizebox{\linewidth}{!}{%
\begin{tikzpicture}[x=1cm,y=1cm]
  \path[use as bounding box] (0,0) rectangle (16,9);
  \draw[draw=panelstroke, fill=panelbg, line width=0.36pt, rounded corners=2pt]
        (0.05,0.05) rectangle (15.95,8.95);

  \fill[white] (0.25,6.05) rectangle (11.35,8.75);
  \draw[draw=panelstroke, line width=0.30pt, rounded corners=2pt] (0.25,6.05) rectangle (11.35,8.75);
  \fill[accentbluepale] (0.25,8.28) rectangle (11.35,8.75);
  \fill[accentblue] (0.25,8.28) rectangle (11.35,8.36);
  \node[anchor=west, font=\sffamily\bfseries\fontsize{6.6pt}{7.0pt}\selectfont, text=accentblue] at (0.55,8.53)
        {Static visual skills};
  \skillmaptile{0.48}{8.07}{3.48}{1.92}{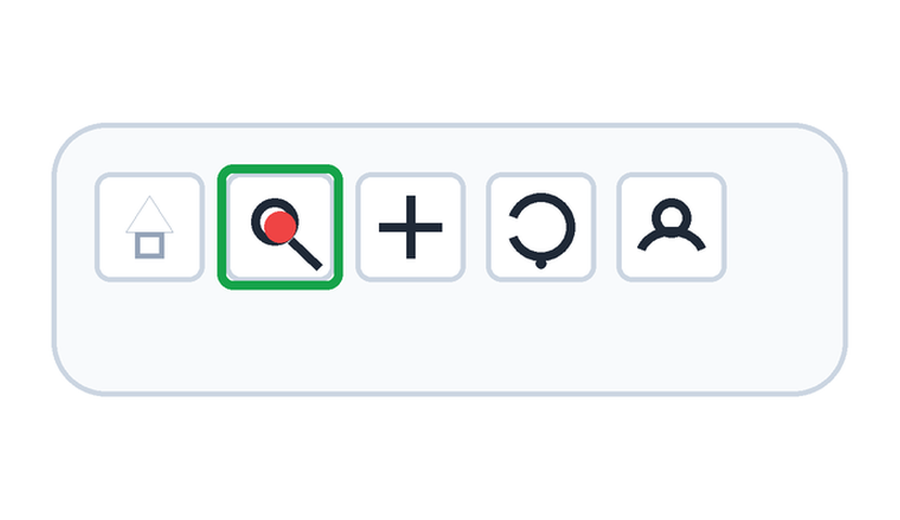}
    {Button click target}
    {Envelope, not icon ink.}{1.32}
  \skillmaptile{4.12}{8.07}{3.48}{1.92}{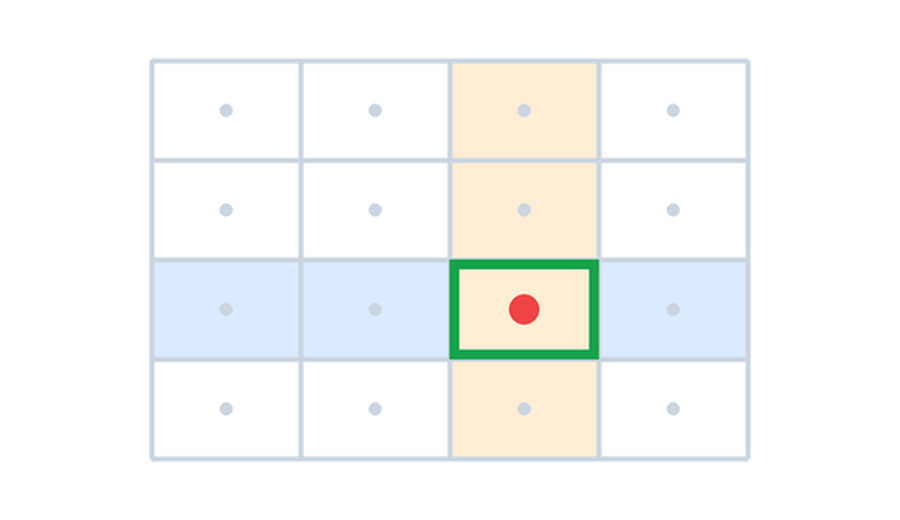}
    {Table cell intersection}
    {Row $\times$ column gives the cell.}{1.32}
  \skillmaptile{7.76}{8.07}{3.48}{1.92}{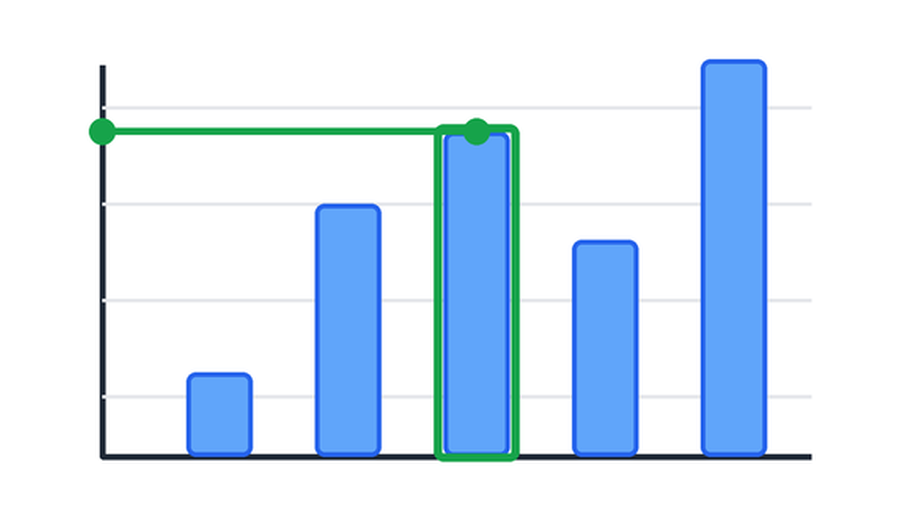}
    {Bar chart projection}
    {Project the mark to the axis.}{1.32}

  \fill[white] (0.25,0.25) rectangle (11.35,5.80);
  \draw[draw=panelstroke, line width=0.30pt, rounded corners=2pt] (0.25,0.25) rectangle (11.35,5.80);
  \fill[accentgreenpale] (0.25,5.34) rectangle (11.35,5.80);
  \fill[accentgreen] (0.25,5.34) rectangle (11.35,5.42);
  \node[anchor=west, font=\sffamily\bfseries\fontsize{6.6pt}{7.0pt}\selectfont, text=accentgreen] at (0.55,5.58)
        {Dynamic visual skills};
  \skillmaptile{0.48}{5.12}{3.48}{2.10}{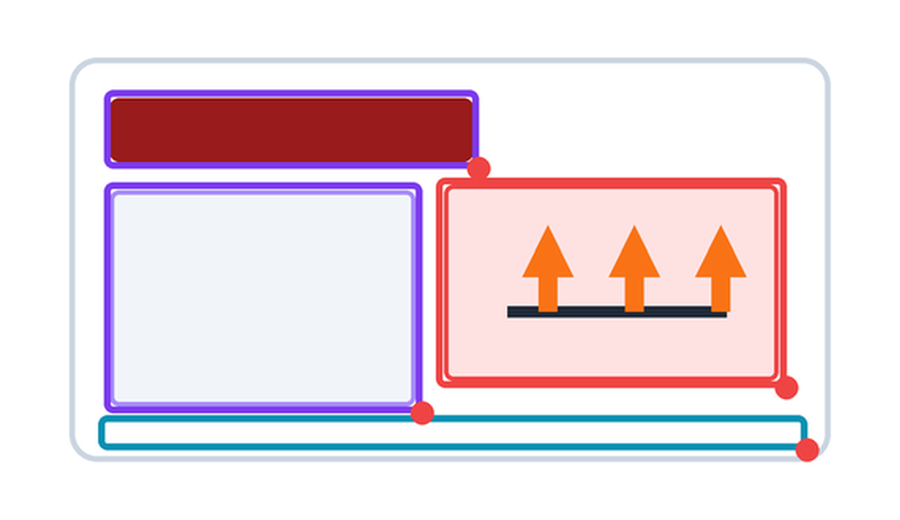}
    {Slide critique}
    {Region marks guide redraw.}{1.46}
  \skillmaptile{4.12}{5.12}{3.48}{2.10}{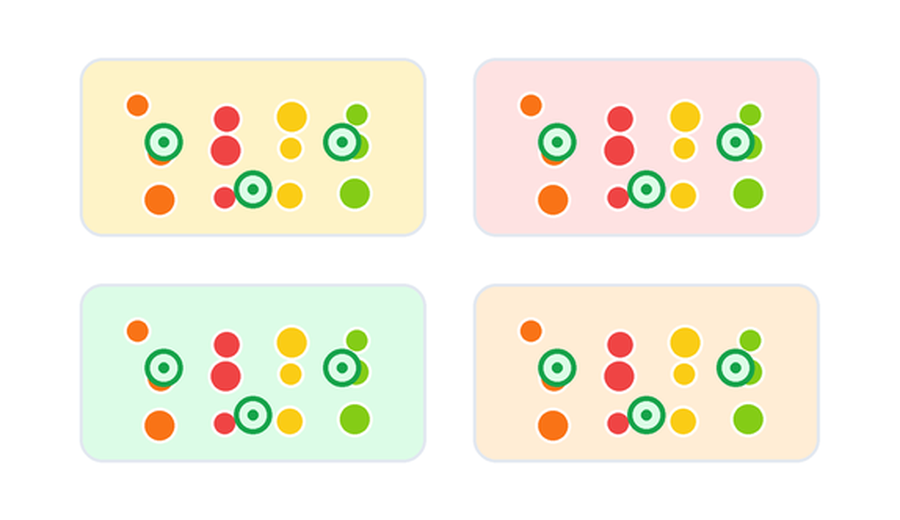}
    {Dense counting}
    {Anchors preserve count state.}{1.46}
  \skillmaptile{7.76}{5.12}{3.48}{2.10}{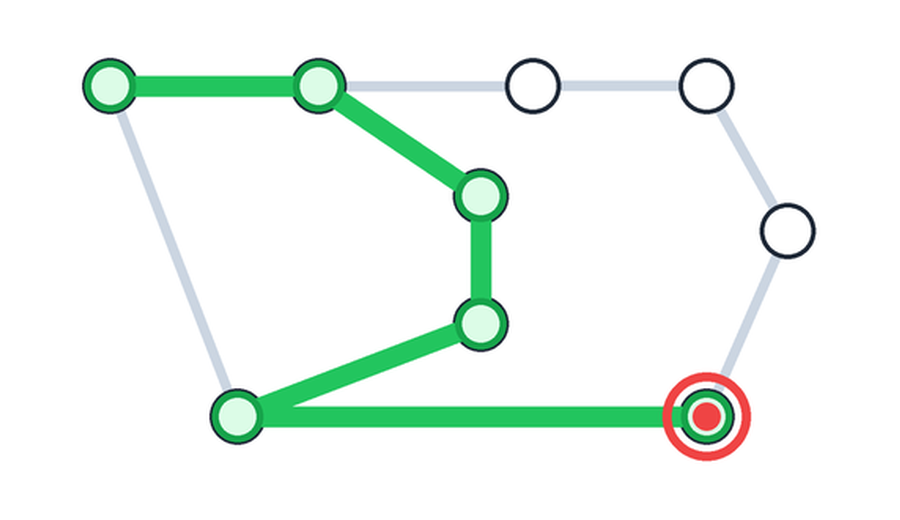}
    {Line tracing}
    {Trace stores visible progress.}{1.46}
  \skillmaptile{0.48}{2.72}{3.48}{2.10}{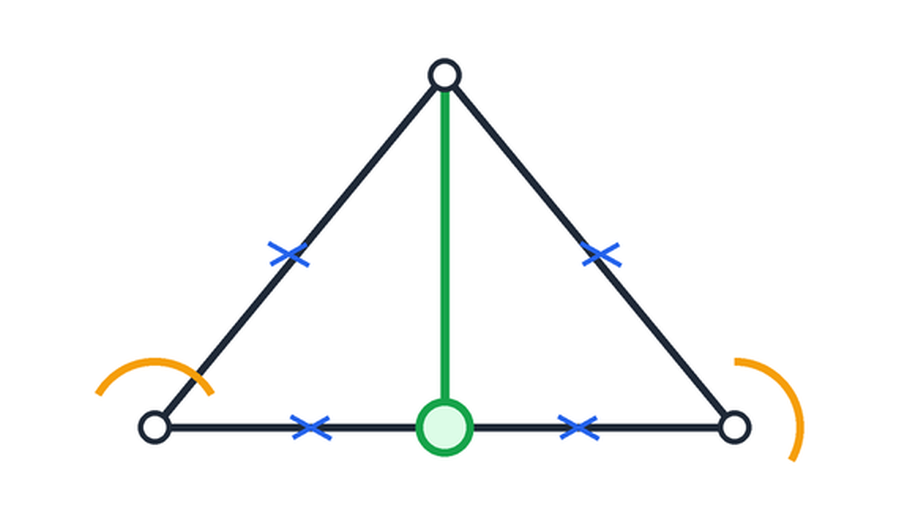}
    {Geometry auxiliary}
    {Auxiliary lines stay on diagram.}{1.46}
  \skillmaptile{4.12}{2.72}{3.48}{2.10}{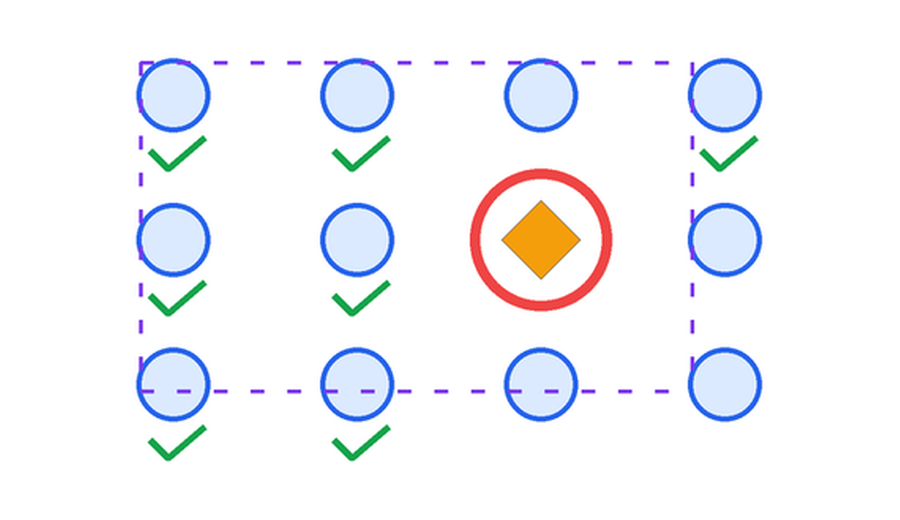}
    {Odd-one-out search}
    {Checked items stay visible.}{1.46}
  \skillmaptile{7.76}{2.72}{3.48}{2.10}{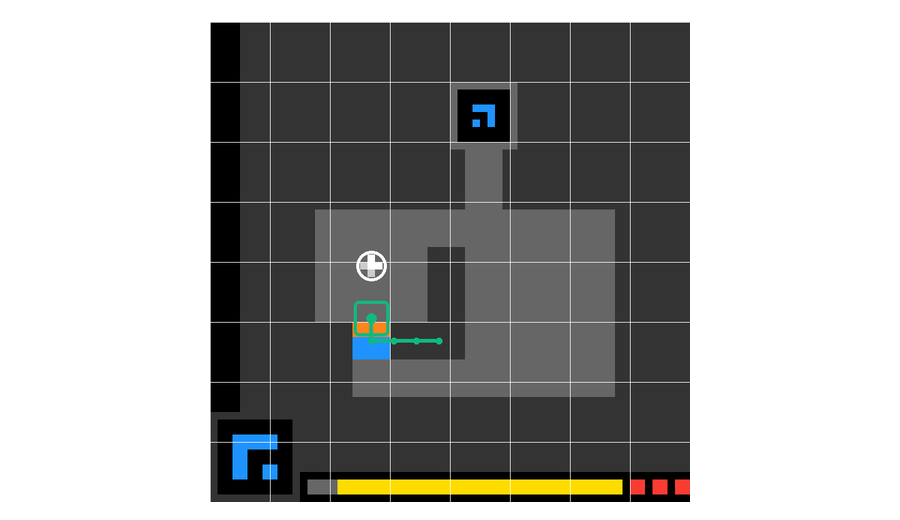}
    {ARC route state}
    {Route memory stays on frame.}{1.46}

  \fill[white] (11.68,0.25) rectangle (15.75,8.75);
  \draw[draw=panelstroke, line width=0.30pt, rounded corners=2pt] (11.68,0.25) rectangle (15.75,8.75);
  \fill[accentorangepale] (11.68,8.28) rectangle (15.75,8.75);
  \fill[accentorange] (11.68,8.28) rectangle (15.75,8.36);
  \node[anchor=west, font=\sffamily\bfseries\fontsize{6.0pt}{6.4pt}\selectfont, text=accentorange] at (11.96,8.53)
        {Interleaved visual skills};
  \skillmaptalltile{11.88}{8.02}{3.64}{2.18}{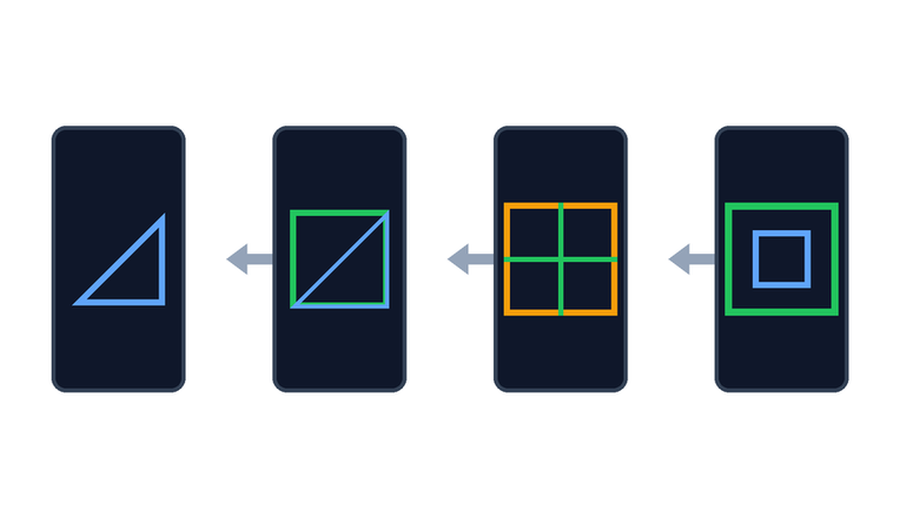}
    {Pythagorean visual proof}
    {Steps stay next to keyframes.}{1.47}
  \skillmaptalltile{11.88}{5.31}{3.64}{2.18}{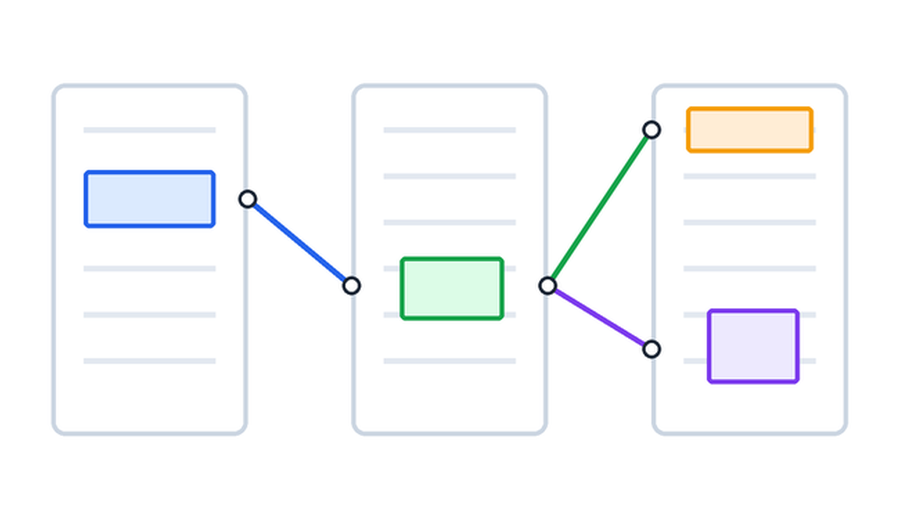}
    {Documentation workflow}
    {Procedures stay next to screenshots.}{1.47}
  \skillmaptalltile{11.88}{2.60}{3.64}{2.18}{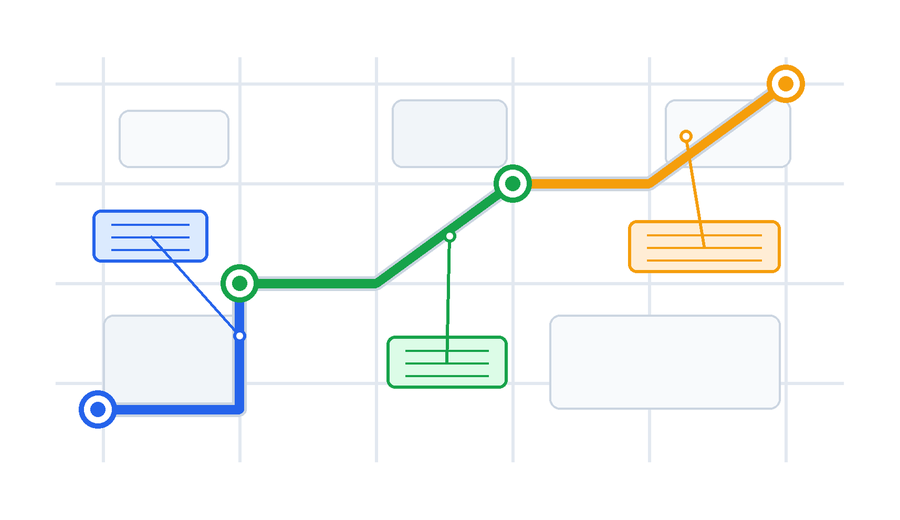}
    {Route-step binding}
    {Directions stay next to map segments.}{1.47}
\end{tikzpicture}%
}
\caption{\textbf{\NAME capability demo.}
Visual skills are organized by the visual bottleneck they solve: static skills clarify reusable spatial conventions; dynamic skills write intermediate state back onto the task image, including slide critique and ARC route-state planning; interleaved skills keep reasoning steps adjacent to their visual evidence.}
\label{fig:skill-capability-map}
\vspace{-3mm}
\end{figure*}

\section{Introduction}

As multimodal large language models (MLLMs)~\cite{achiam2023gpt,bai2023qwen,team2023gemini,hua2025v2xum,liu2024deepseek,sun2026latent} increasingly serve as the reasoning backbone of general-purpose multimodal agents~\cite{gu2025ui,zhou2025mai,ye2025mobile,zeng2025mira}, reusable skills and tools have become a key mechanism for scaling agent capabilities to complex long-horizon tasks~\cite{YaoEtAl2023,SchickEtAl2023,huang2025building,li2026jobbench}. 
Recent systems accumulate and reuse agent experience in the form of prompt-based standard operating procedures, tool-use interfaces, workflow templates, and skill libraries~\cite{WuEtAl2023,LangChain2026}. 
However, existing skill construction largely relies on purely textual abstractions of task experience: demonstrations or past interactions are typically distilled into structured natural language that specifies goals, action sequences, input-output formats, and exception-handling rules. Such abstractions are particularly effective for symbolic tasks such as API configuration, database querying, and logical reasoning, where reusable knowledge is naturally procedural and linguistic in nature~\cite{SchickEtAl2023}.

For visual-centric tasks~\cite{thrush2022winoground,hsieh2023sugarcrepe,hua2024finematch,hua2024mmcomposition}, however, purely textual skills are inherently limited. 
GUI manipulation requires skills to encode not only what action to take, but also where and how the action should be grounded, including the visual extent of controls, icon hit regions, hierarchical nesting, and nearby distractors~\cite{cheng-etal-2024-seeclick}. 
Similarly, UI design and poster generation depend on reusable visual patterns such as module proportions, whitespace rhythm, spatial alignment, and visual hierarchy~\cite{SiEtAl2025,HsuEtAl2023}. 
Counting, maze solving, and other visual verification tasks further require spatial traversal strategies, localized inspection routines, and consistency-checking protocols~\cite{GuptaKembhavi2023}. 
While these factors can be partially verbalized, compressing high-dimensional visual procedures into text can discard spatial evidence and introduce ambiguity in execution~\cite{LarkinSimon1987Diagram}. 
This reveals a textual bottleneck in current skill construction: for multimodal agents, reusable skills should preserve not only procedural instructions, but also the visual priors needed to ground, inspect, and verify actions.

This bottleneck is already evident in existing visual-agent systems. 
WebArena~\cite{ZhouEtAl2023} and Mind2Web~\cite{DengEtAl2023} cast web operation as mapping natural-language goals to action sequences, yet strong models remain far below human performance on long-horizon tasks~\cite{ZhouEtAl2023,DengEtAl2023,zeng2026mementogui}. 
In mobile control, AppAgent still requires models to rediscover actionable regions from each new screenshot~\cite{ZhangEtAl2023}, while GUI grounding benchmarks such as SeeClick and ScreenSpot consistently show that knowing \enquote{what to click} does not necessarily imply knowing \enquote{where to click} with sufficient spatial precision~\cite{cheng-etal-2024-seeclick,yu2026aurora}. Similarly, layout generation benchmarks such as Design2Code, WebSight, and PosterLayout show that translating visual structures into code or layouts requires reusable visual conventions over proportion, spacing, alignment, and hierarchy, beyond recognizing textual content~\cite{SiEtAl2025,hu2023promptcap,LaurenconEtAl2024,HsuEtAl2023,hua2024finematch,Hua2025MMIGBenchTC}. 
Taken together, these findings motivate the design of reusable skills that explicitly preserve task-relevant spatial evidence and visual regularities, rather than relying solely on textualized procedures.

In this paper, we frame this limitation as a textual bottleneck in reusable skill construction. 
The key issue is not that multimodal agents cannot perceive images, but that their accumulated task experience is often stored and reused through text-dominated skill representations. 
As a result, a skill may describe what to do while failing to preserve the visual traces needed to guide where to look, how to inspect, and how to verify outcomes. 
We further observe that different visual tasks inherently require different forms of visual support. Tasks governed by stable spatial conventions, such as GUI grounding, particularly benefit from static visual references that encode reusable interaction protocols. In contrast, tasks requiring continuous perceptual tracking, such as dense counting, benefit from dynamic \textit{in-situ} traces that effectively externalize intermediate spatial state during reasoning. A third family of tasks, such as document workflows, visual tutorials, and evidence-grounded explanations, is best represented by interleaving each textual step with the visual source evidence it depends on, so that the reusable skill remains grounded in ordered frames, screenshots, or page regions rather than becoming a detached prose summary.

To address this issue, we propose \textbf{\NAME}, a new agent-skill paradigm that extends conventional text-only skills into reusable multimodal entities. 
Under this paradigm, a skill is no longer merely a textual instruction or reasoning trace, but a multimodal asset composed of three complementary components. 
First, declarative textual logic coordinates semantic reasoning, execution steps, and task-specific boundary conditions. Second, visual priors and source-grounded references encode reusable spatial topology, visual boundaries, commonly observed error patterns, and step-specific evidence across diverse instances. Third, a multimodal binding protocol explicitly specifies how textual logic and visual support should be jointly grounded, retrieved, and executed during the course of task solving. As illustrated in~\cref{fig:skill-capability-map}, \NAME preserves visual procedural knowledge as reusable multimodal content, enabling agents to seamlessly combine textual control with explicit visual guidance for spatially intensive tasks.

The core contributions of this paper are as follows:

\begin{enumerate}
    \item \textbf{Identifying the textual bottleneck.} 
    We analyze a key limitation of current skill-learning paradigms: text-only skills are poorly suited to preserving spatial structure, visual boundaries, and perceptual tracking protocols required by visual-centric agent tasks.

    \item \textbf{Formulating and instantiating visual skills.} 
    We introduce \NAME, a reusable multimodal skill representation that combines declarative textual logic, visual priors or references, and a multimodal binding protocol. 
    We define static, dynamic, and interleaved visual skills as three complementary ways of packaging reusable visual support.
    We further develop \SYSTEM, a proof-of-concept authoring pipeline that automatically diagnoses visual bottlenecks, generates textual and visual skill components, and packages them into reusable \NAME artifacts.

    \item \textbf{Validating visual priors across tasks.} Through controlled experiments on two representative tasks, GUI grounding and dense object counting, we show that both static visual references and dynamic \textit{in-situ} traces consistently improve over text-only skills, highlighting visual structure as a first-class and previously underexplored asset for multimodal agents.

\end{enumerate}

\section{The Textual Bottleneck in Current Skill Paradigms}
\label{sec:bottleneck}

The textual bottleneck reflects a fundamental mismatch between what multimodal agents perceive and what text-only skills preserve. 
Although MLLMs can process visual inputs, their reusable skills are often stored as textual prompts, reasoning traces, or summarized trajectories. 
This creates a representational gap for visual interfaces, where task-relevant knowledge is organized as continuous spatial signals: controls occupy precise regions, modules obey proportional relationships, and elements interact through alignment, occlusion, and connectivity. 
Compressing such visual interaction protocols into text reduces high-dimensional spatial topology to a one-dimensional symbolic sequence~\cite{TishbyEtAl1999,Shannon1959RateDistortion}. 
Because diagrammatic and sentential representations support different perceptual inferences~\cite{LarkinSimon1987Diagram}, this compression can lose spatial evidence that is difficult to recover through verbal detail alone.

Motivated by our empirical observations across visual-agent tasks, we identify two recurring failure modes induced by this representational gap.
The first is \textbf{static protocol ambiguity}.
Many visual actions depend on fine-grained cues such as boundary curvature, relative displacement, hit-region tolerance, whitespace rhythm, visual hierarchy, and layout proportion.
While an MLLM may perceive these cues in a screenshot, a text-only skill cannot reliably preserve them as reusable procedural knowledge: once translated into language, they become underspecified or detached from their original spatial context~\cite{LarkinSimon1987Diagram,Ainsworth2006DeFT,Paivio1986,Vessey1991}.
Thus, text-only skills may describe what to do but fundamentally fail to encode the visual conventions needed to execute it reliably in practice. Moreover, adding more textual boundary conditions only partially mitigates this underlying issue, often increasing brittleness and reasoning burden without restoring the task's native spatial structure~\cite{Sweller1988CognitiveLoad,ChandlerSweller1991Format,SwellerEtAl1998CognitiveArchitecture,MayerMoreno2003CognitiveLoad}.

The second failure mode is \textbf{dynamic tracking collapse}. Dense counting, maze solving, and spatial verification all require a persistent record of which regions have been inspected, which instances have already been counted, and where attention should move next. Text-only traces can store this state only as coordinate lists or verbal descriptions, which rapidly become ambiguous as visual density increases. This inevitably leads to omissions, repeated inspections, and double-counting, not because the agent lacks the task rule, but because text is inherently a poor medium for maintaining continuous spatial bookkeeping~\cite{LarkinSimon1987Diagram}. Such tasks fundamentally require skills that support visually grounded intermediate state tracking, rather than relying on procedural instructions alone.

These two failure modes share the same root: text-only skills do not preserve visual structure as reusable procedural knowledge. 
However, they call for different forms of visual support. 
For tasks governed by stable spatial conventions, reusable knowledge should be stored as visual priors rather than expanded into increasingly complex textual rules. 
For tasks requiring continuous perceptual bookkeeping, intermediate reasoning should be externalized as in-situ visual traces that remain grounded in the task image. 
Together, these observations motivate our core design principle: reusable agent skills should go beyond text by treating visual structure as a first-class skill asset.

\section{\NAME: Visual Structure as a First-Class Skill Asset}
\label{sec:visual_skill}

\subsection{Definition}

We define \NAME as a reusable multimodal skill entity for agents. 
Unlike text-only skills that store experience as instructions, reasoning traces, or summarized trajectories, \NAME represents a skill as
\[
\mathcal{S}_{v} = (\mathcal{L}, \mathcal{P}_{v}, \mathcal{B}),
\]
where $\mathcal{L}$ denotes declarative textual logic, $\mathcal{P}_{v}$ denotes reusable visual priors or source-grounded visual references, and $\mathcal{B}$ denotes the binding protocol that governs their joint execution.

\begin{enumerate}
    \item \textbf{Declarative textual logic.} 
    $\mathcal{L}$ specifies the task objective, execution procedure, input-output constraints, boundary conditions, and corresponding failure-handling strategies. It fully preserves the well-established strengths of conventional text-based skills, including abstraction, compositionality, interpretability, and procedural control over task execution.
    
    \item \textbf{Reusable visual support.} 
    $\mathcal{P}_{v}$ preserves task-relevant visual structure that is difficult to encode in text. 
    We instantiate three forms of visual support to address different visual bottlenecks:
    \begin{itemize}
    \item \textbf{Static priors} are external visual references, such as wireframes, layout prototypes, annotation templates, or error-pattern examples. They capture stable spatial conventions shared across task instances and mitigate \textit{static protocol ambiguity} by providing reusable references for layout, boundary, alignment, and interaction patterns.

    \item \textbf{Dynamic priors} are executable spatial protocols for \textit{in-situ} visual tracking during inference. Rather than storing fixed images, they specify how to initialize, update, and verify intermediate visual traces, such as anchors, trajectories, visited regions, or counting marks. They mitigate \textit{dynamic tracking collapse} by turning spatial bookkeeping into a grounded and continuously maintained visual working-memory process.

    \item \textbf{Interleaved visual skills} bind ordered textual steps to the source evidence that supports them, such as video keyframes, documentation screenshots, page regions, or source-image crops. They are useful when the reusable knowledge is not a single prior, but a step-to-evidence structure that keeps procedural language visually grounded.
    \end{itemize}

    \item \textbf{Multimodal binding protocol.} 
    $\mathcal{B}$ specifies when and how textual logic should be grounded with visual priors. 
    For each reasoning step, it determines whether to retrieve a static prior, instantiate a dynamic prior, bind an interleaved source reference, or proceed with text alone. It also prevents static references from being mistaken for task instances, standardizes dynamic-trace updates, and keeps interleaved claims adjacent to their grounding evidence.
\end{enumerate}

\begin{algorithm}[htbp]
    \caption{Multimodal Binding Protocol}
    \label{alg:binding_logic}
    \begin{algorithmic}[1]
        \For{each reasoning step $s_i$}
            \State $p_i \leftarrow \varnothing$
            \If{$s_i$ requires visual support}
                \If{$s_i$ depends on stable spatial conventions}
                    \State $p_i \leftarrow \textsc{RetrieveStaticPrior}(s_i)$
                \ElsIf{$s_i$ requires \textit{in-situ} spatial tracking}
                    \State $p_i \leftarrow \textsc{InstantiateDynamicPrior}(s_i)$
                \ElsIf{$s_i$ depends on ordered source evidence}
                    \State $p_i \leftarrow \textsc{BindInterleavedReference}(s_i)$
                \EndIf
                \State Bind $s_i$ and $p_i$ under role-specific constraints
            \EndIf
            \State Execute $s_i$ on the task input, guided by $p_i$ if bound
            \If{$p_i$ is a dynamic prior}
                \State Update $p_i$ with new spatial outputs
            \EndIf
        \EndFor
    \end{algorithmic}
\end{algorithm}

Thus, \NAME extends text-based skills from linguistic procedures to multimodal skill assets: textual logic specifies what to do, visual priors and references preserve where and how to inspect, and the binding protocol determines when these components should jointly guide execution.

\subsection{Separation of Responsibilities: Text for Logic, Vision for Space}

\NAME clarifies the division of labor between modalities rather than reducing the role of text. 
Text is well-suited for specifying \emph{what to do}: parsing instructions, organizing execution steps, resolving semantic ambiguity, and defining output formats. 
Vision is better suited for preserving \emph{where and how to inspect}: hit regions, layout proportions, icon-versus-container granularity, counting scan order, and spatial violation patterns. 
This separation is not an absolute dichotomy; text can describe coarse relations such as \enquote{A is above B}. 
However, when spatial information becomes dense, continuous, or geometrically precise, representing it only in language introduces an information bottleneck. Delegating complex spatial grounding to visual priors keeps textual logic concise and composable, while making spatial knowledge reusable, inspectable, and refinable across instances. 
A visual prior therefore serves as task-level spatial knowledge: a reusable map of where and how to look, rather than a long verbal description of the same structure.

\begin{figure}[t]
\centering
\includegraphics[width=\linewidth]{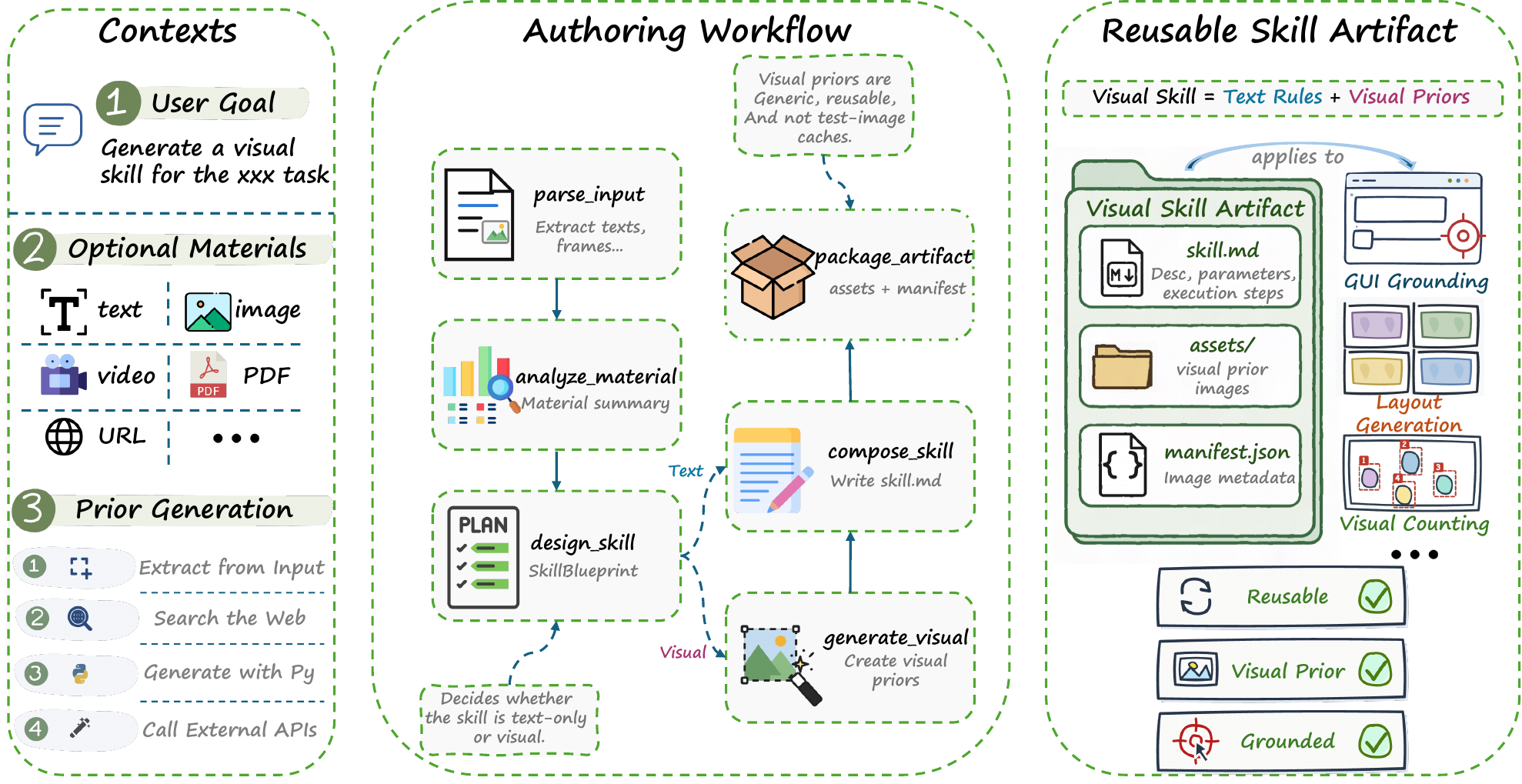}
\caption{\textbf{Authoring reusable visual skills from multimodal context.}
\SYSTEM converts a user goal and optional multimodal materials into a reusable visual skill by analyzing task constraints, generating visual priors or source-grounded references when needed, and packaging them with textual logic and binding manifests for cross-instance transfer.}
\label{fig:arch}
\vspace{-4mm}
\end{figure}
    
\section{Paradigm Instantiation: The \SYSTEM Framework}
\label{sec:autoskill}

To instantiate visual skills in a scalable and reproducible manner, we introduce \textbf{\SYSTEM}, a proof-of-concept authoring pipeline that synthesizes \textbf{\NAME} artifacts from user goals and multimodal context using foundation-model APIs. Each run produces a self-contained skill directory with \texttt{skill.md}, \texttt{manifest.json}, visual assets, and provenance records, so the artifact can be loaded by an agent, inspected by a human, or versioned in a repository (\cref{fig:arch}). The skills used in our empirical study are all generated by \SYSTEM, demonstrating cross-environment transfer without task-specific manual skill authoring.

\paragraph{Input and normalization.}
\SYSTEM takes a user goal and optional multimodal context, such as text, images, accessible URLs, and sampled video frames. It normalizes them into semantic text, visual frames, and metadata, retrieves supplementary domain knowledge when needed, and extracts task constraints plus candidate reusable visual protocols.

\paragraph{Visual-bottleneck gate.}
A diagnostic gate decides whether a task can remain text-only or needs visual support. It checks whether the task requires spatial grounding, structural geometry, perceptual tracking, or reusable source evidence, and whether the proposed visual support encodes a cross-instance protocol rather than a few-shot image cache. The gate also separates the public \emph{skill kind} from the lower-level prior mechanism: static and dynamic skills instantiate the evaluated prior families, while interleaved skills organize ordered text--visual evidence bindings.

\paragraph{Dual-track generation.}
Generation proceeds along two tracks. The linguistic track writes declarative logic, while the visual track extracts source regions, retrieves missing conventions, renders diagrams or dynamic overlays, or invokes generative visual models. The components are then packaged with binding manifests into reusable \NAME artifacts.

\paragraph{Execution-oriented artifacts.}
The manifest records the skill kind, prior kind, asset roles, renderer strategy, binding rules, and usage constraints. This lets downstream agents choose whether to load a fixed prior, maintain an iterative visual-state loop, or present source evidence adjacent to the corresponding reasoning step.

\paragraph{Open-source interface.}
The released system provides a command-line interface, a lightweight Gradio demo, example skills, and minimal agent-integration code. Additional examples across all three skill forms are shown in \cref{app:example-gallery} and released with the project repository.

\section{Empirical Study}

\subsection{Controlled Settings}
To evaluate the necessity of visual protocols and validate our taxonomy of cognitive bottlenecks, we use three unified settings across all tasks:

\begin{enumerate}
    \item \textbf{No-skill setting (Direct Prompting).} The model receives only the original task input image and the user query, without any additional skill intervention.

    \item \textbf{Text-only skill setting.} The model receives the frozen declarative textual logic, detailing the execution steps and output format, but does not receive any explicit spatial priors.

    \item \textbf{Visual-skill setting.} The model receives the same declarative textual logic as in the text-only setting. However, depending on the cognitive bottleneck of the specific task, it is additionally equipped with an explicit spatial prior---implemented either as a \textbf{Static Prior} (an external reference diagram) or a \textbf{Dynamic Prior} (textual rules forcing \textit{in-situ} coordinate generation).
\end{enumerate}
It is important to emphasize that the goal of our empirical study is not to pursue a new state of the art on these benchmarks, but rather to measure the textual degradation rate. The text-only and visual-skill settings use the exact same set of foundational textual rules throughout all experiments. The only variable is whether a visual prior (in its task-appropriate form) is introduced alongside the textual logic. Therefore, the performance gain $\Delta$ between the two settings can be interpreted as the value of spatial information that text-only skills fail to encode but visual protocols can successfully recover.

\paragraph{Why we do not isolate interleaved skills as a third benchmark condition.}
Interleaved visual skills are a packaging and binding form rather than a third atomic prior. Their value comes from keeping ordered reasoning adjacent to source-grounded evidence, while the underlying operations still rely on the two primitive bottlenecks evaluated here: static grounding of regions and dynamic maintenance of visual state. A separate benchmark would conflate source quality, frame selection, document layout, and scoring. We therefore evaluate the primitive mechanisms under controlled settings and illustrate interleaved skills through examples in \cref{app:example-gallery}.

\subsection{Tasks and Models}

\NAME covers many examples across static, dynamic, and interleaved forms; representative cases are shown in the appendix and the project repository. For controlled empirical evidence, we select two canonical tasks that isolate the two primitive mechanisms: GUI grounding for static spatial conventions, and dense object counting for dynamic visual working memory. We pair each task with a strong foundation model to stress-test whether \textsc{Visual Skill} remains useful atop favorable baselines.

\begin{enumerate}
    \item \textbf{GUI Grounding (Static Priors).} The model localizes interaction targets from screenshots, evaluated on ScreenSpot~\cite{cheng-etal-2024-seeclick}, ScreenSpot-v2~\cite{wu2024osatlas}, and GroundUI-18K~\cite{zheng2024agentstudio} with Point-in-Box Accuracy as the primary metric. We use \textbf{Qwen3-VL-32B-Thinking}~\cite{bai2025qwen3vl}.

    \item \textbf{Dense Object Counting (Dynamic Priors).} On CountBenchQA~\cite{beyer2024paligemma,paiss2023countclip}, the agent performs iterative spatial enumeration by anchoring one coordinate point per valid instance. Metrics include exact-match Accuracy, MAE, and Within-1 Accuracy. We use \textbf{Gemini-2.5-Pro}~\cite{comanici2025gemini25}.
\end{enumerate}

Results show that, regardless of each model's inherent specialization, purely textual skill descriptions consistently suppress its full potential, whereas our \textbf{\NAME} yields significant and consistent improvements across all evaluated benchmarks and associated metrics.

\subsection{Measuring textual degradation in reusable agent skills.}
Beyond proposing visual skills as a richer skill representation, we argue that the limitation of text-only skills should be made measurable. We define \emph{textual degradation} as the performance loss incurred when reusable task knowledge is forced to be encoded purely in text, while keeping the underlying task rules and agent backbone unchanged. Formally, for a task family $\mathcal{T}$ and a performance metric $M$, we measure
\begin{equation}
\mathrm{TDR}(\mathcal{T}) =
M(\pi_{\mathrm{visual\ skill}}, \mathcal{T}) -
M(\pi_{\mathrm{text\ skill}}, \mathcal{T}),
\end{equation}
where $\pi_{\mathrm{text\ skill}}$ and $\pi_{\mathrm{visual\ skill}}$ use the same textual task rules, but the latter additionally has access to reusable visual priors such as spatial layouts, region bindings, appearance prototypes, trajectory snippets, or visual verification protocols. To compare across task families, we further define a normalized textual degradation rate:
\begin{equation}
\mathrm{nTDR}(\mathcal{T}) =
\frac{
M(\pi_{\mathrm{visual\ skill}}, \mathcal{T}) -
M(\pi_{\mathrm{text\ skill}}, \mathcal{T})
}{
M(\pi_{\mathrm{oracle\ visual}}, \mathcal{T}) -
M(\pi_{\mathrm{text\ skill}}, \mathcal{T})
}.
\end{equation}
This metric quantifies how much recoverable task-relevant information is lost under text-only skillization. A high TDR indicates that the task depends on reusable knowledge that is difficult to faithfully serialize into language, such as geometry, localized visual evidence, object identity, layout constraints, or perceptual decision criteria. TDR can be instantiated with task-specific metrics, including point-in-box accuracy and center-distance for GUI grounding, exact accuracy or MAE for dense counting, evidence-region matching for document workflows, alignment and overlap metrics for layout generation, route-step binding accuracy for map navigation, and omission, duplication, or error-localization rates for visual verification. In this sense, TDR turns the ``textual bottleneck'' from a qualitative claim into an evaluation target: it asks which forms of reusable agent knowledge can be safely textualized, which suffer the largest degradation, and which visual priors are most valuable to preserve.

\subsection{Experimental Results and Analysis}

\begin{table*}[t]
  \centering
  \small
  \setlength{\tabcolsep}{7pt}
  \renewcommand{\arraystretch}{1.12}
  \caption{\textbf{Evaluating Static Priors on Protocol Ambiguity.}
  We compare no-skill, text-only skills, and Visual Skills (equipped with Static Priors) across three GUI grounding benchmarks.
  Across all GUI icon samples, Visual Skill significantly improves Point-in-Box accuracy over No-skill (91.1\% vs. 86.4\%, $p=0.005$) and shows a positive trend over Text-only Skill (91.1\% vs. 88.1\%, $p=0.067$).}
  \label{tab:static-prior-results}
  \begin{tabular}{l|lccc}
    \toprule
    \textbf{Benchmark} & \textbf{Method}
    & \textbf{Point-in-Box Acc} $\uparrow$
    & \textbf{Mean IoU} $\uparrow$
    & \textbf{Mean Center Dist} $\downarrow$ \\
    \midrule
    \multirow{3}{*}{ScreenSpot}
      & No-skill & 0.873 & 0.274 & 0.037\\
      & Text-only Skill  & 0.901 & 0.318 &  0.035\\
      & \cellcolor{violet!10}\textbf{Visual Skill}
      & \cellcolor{violet!10}\textbf{0.930}
      & \cellcolor{violet!10}\textbf{0.364}
      & \cellcolor{violet!10}\textbf{0.030} \\
    \midrule
    \multirow{3}{*}{ScreenSpot-v2}
      & No-skill  & 0.917 & 0.307& 0.022  \\
      & Text-only Skill & 0.923 & 0.343 & 0.021 \\
      & \cellcolor{violet!10}\textbf{Visual Skill}
      & \cellcolor{violet!10}\textbf{0.951}
      & \cellcolor{violet!10}\textbf{0.418}
      & \cellcolor{violet!10}\textbf{0.019} \\
    \midrule
    \multirow{3}{*}{GroundUI-18K}
      & No-skill  & 0.670 & 0.317 & 0.075 \\
      & Text-only Skill& 0.686 & 0.335 & 0.074 \\
      & \cellcolor{violet!10}\textbf{Visual Skill}
      & \cellcolor{violet!10}\textbf{0.713}
      & \cellcolor{violet!10}\textbf{0.376}
      & \cellcolor{violet!10}\textbf{0.069} \\
    \bottomrule
  \end{tabular}
  \vspace{-4mm}
\end{table*}

\subsubsection{Static Priors: Resolving Protocol Ambiguity through Visual References}

We evaluate static priors on GUI grounding, where successful execution depends on implicit interaction conventions and boundary-sensitive localization. As shown in \cref{tab:static-prior-results}, the no-skill baseline achieves reasonable Point-in-Box Accuracy but remains notably less precise on boundary-sensitive metrics such as Mean IoU and Mean Center Distance. Adding text-only skills improves performance in some cases, suggesting that declarative procedures can help the model parse instructions and filter candidate targets more systematically. However, these gains remain limited, particularly for metrics that require spatially accurate localization of fine-grained functional UI regions.

In contrast, equipping the same textual logic with static visual priors yields consistent and measurable improvements across all three GUI grounding benchmarks. The largest gains appear in Mean IoU, indicating that visual references help calibrate the model's understanding of target extent, hit regions, and fine-grained UI boundaries more effectively. These results support our hypothesis that many GUI grounding errors arise not only from missing procedural rules, but also from fundamentally underspecified spatial conventions that are difficult to adequately encode in text alone.

Instantiating TDR on GUI grounding by averaging the three benchmarks gives a textual-degradation gap of $+0.028$ Point-in-Box accuracy, $+0.054$ Mean IoU, and a $0.0040$ Mean Center Distance reduction, corresponding to normalized degradation rates of $17.1\%$, $8.1\%$, and $9.2\%$, respectively.

This performance pattern illustrates the role of static priors as reusable visual dictionaries. 
Text-only skills can describe how to reason about target categories, candidate filtering, and output format, but they do not directly preserve the spatial regularities of UI elements, such as nested controls, dense toolbars, implicit hitboxes, and tolerance regions. 
Static priors complement textual logic by providing explicit visual references for these conventions, allowing the agent to ground abstract interaction rules in pixel-level structure. 
Rather than replacing textual reasoning, static priors provide modality-matched spatial guidance for boundary-sensitive GUI execution.

\subsubsection{Dynamic Priors: Supporting Perceptual Tracking through \textit{In-situ} Generation}

We evaluate dynamic priors on dense spatial reasoning tasks such as object counting, where the key challenge is maintaining a persistent record of inspected regions and counted instances. 
In these settings, text-only rules can instruct the model to count carefully or avoid duplicates, but they do not provide an explicit spatial memory for tracking which objects have already been visited. 
As visual density increases, this can lead to omissions, repeated inspections, or inconsistent enumeration.

The results on CountBenchQA in \cref{tab:dynamic-prior-results} show that text-only skills do not improve over direct prompting and can even reduce performance. 
This suggests that additional procedural instructions may introduce reasoning overhead when they are not paired with a grounded mechanism for spatial bookkeeping. 
By contrast, dynamic priors improve accuracy and substantially reduce MAE, indicating that explicit spatial anchoring helps the model maintain a more consistent counting state.

Dynamic priors address this bottleneck by externalizing intermediate state as an \textit{in-situ} visual trace. 
Instead of relying only on an internal textual chain of thought, the model iteratively plots coordinate anchors on target objects and receives the rendered anchors as visual working memory. 
This shifts counting from implicit global estimation to explicit spatial enumeration, making the model's intermediate decisions inspectable and reducing duplicate or missed counts. 
These results support our claim that tasks requiring continuous spatial bookkeeping benefit from visual traces that remain grounded in the task image, rather than from procedural text alone.

% \FloatBarrier
% \clearpage

\begin{table*}[htbp]
  \centering
  \small
  \setlength{\tabcolsep}{12pt}
  \renewcommand{\arraystretch}{1.12}
  \caption{\textbf{Evaluating Dynamic Priors on Dense Perception.}
  Results on CountBenchQA. Pure textual rules (\textbf{Text-only Skill}) increase error variance, whereas generating an \textit{in-situ} spatial trajectory (\textbf{Visual Skill}) boosts exact-match accuracy and reduces MAE by $\sim$60\% over the baseline.
  The improvement is statistically significant: Visual Skill outperforms Text-only Skill by $+4.12$ points in exact accuracy ($p=0.003$) and No-skill by $+2.88$ points ($p=0.027$).}
  \label{tab:dynamic-prior-results}
  \begin{tabular}{l|lccc}
    \toprule
    \textbf{Benchmark} & \textbf{Method}
    & \textbf{Acc. (\%)} $\uparrow$
    & \textbf{MAE} $\downarrow$
    & \textbf{Within-1 Acc. (\%)} $\uparrow$ \\
    \midrule
    \multirow{3}{*}{CountBenchQA}
      & No-skill
      & 94.24
      & 0.1317
      & 97.74 \\
      & Text-only Skill
      & 93.00
      & 0.1612
      & 96.30 \\
      & \cellcolor{violet!10}\textbf{Visual Skill}
      & \cellcolor{violet!10}\textbf{97.12}
      & \cellcolor{violet!10}\textbf{0.0535}
      & \cellcolor{violet!10}\textbf{98.97} \\
    \bottomrule
  \end{tabular}
  \vspace{-4mm}
\end{table*}

Instantiating TDR on CountBenchQA gives a textual-degradation gap of $+4.12$ exact-accuracy points, $+2.67$ Within-1 accuracy points, and a $0.1077$ MAE reduction from text-only skills to visual skills, corresponding to normalized degradation rates of $58.9\%$, $72.2\%$, and $66.8\%$, respectively.

\begin{figure*}[t]
  \centering
  \scriptsize
  \setlength{\fboxsep}{5pt}
  \definecolor{VSqualGuiHead}{HTML}{FAD1D1}
  \definecolor{VSqualGuiBg}{HTML}{FFF3F3}
  \definecolor{VSqualCountHead}{HTML}{FFE2B8}
  \definecolor{VSqualCountBg}{HTML}{FFF8ED}
  \definecolor{VSqualQuestion}{HTML}{FFF4C2}
  \definecolor{VSqualGreen}{HTML}{078B63}
  \definecolor{VSqualMuted}{HTML}{6B7280}

  \newcommand{\qualcaption}[3]{%
    \vspace{1pt}
    \colorbox{VSqualQuestion}{\parbox{0.94\linewidth}{\centering\textit{#1}}}\\[-1pt]
    \parbox{0.96\linewidth}{\centering #2; T #3; {\color{VSqualGreen}\textbf{V}}}
  }
  \newcommand{\qualcaptionfull}[4]{%
    \vspace{1pt}
    \colorbox{VSqualQuestion}{\parbox{0.94\linewidth}{\centering\textit{#1}}}\\[-1pt]
    \parbox{0.96\linewidth}{\centering #2; T #3; {\color{VSqualGreen}\textbf{V #4}}}
  }

  \noindent\colorbox{VSqualGuiHead}{%
    \parbox{\dimexpr\textwidth-2\fboxsep\relax}{%
      \normalsize\textbf{GUI Grounding Cases}
      \hfill {\color{VSqualMuted}visual priors improve hitbox precision}
    }}
  \vspace{2pt}

  \noindent\colorbox{VSqualGuiBg}{%
    \parbox{\dimexpr\textwidth-2\fboxsep\relax}{%
      \begin{minipage}[t]{0.238\linewidth}
        \centering
        \includegraphics[width=\linewidth]{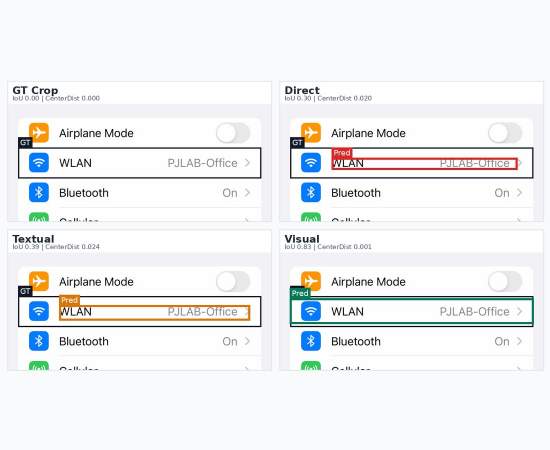}
        \qualcaptionfull{``check wlan settings''}{D 0.30}{0.39}{0.83}
      \end{minipage}\hfill
      \begin{minipage}[t]{0.238\linewidth}
        \centering
        \includegraphics[width=\linewidth]{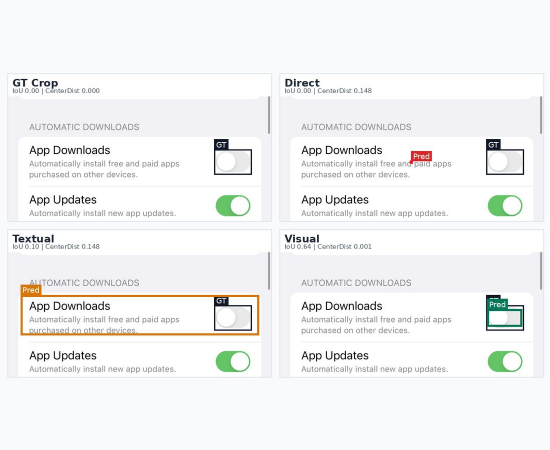}
        \qualcaptionfull{``open app automatic download''}{D 0.00}{0.10}{0.64}
      \end{minipage}\hfill
      \begin{minipage}[t]{0.238\linewidth}
        \centering
        \includegraphics[width=\linewidth]{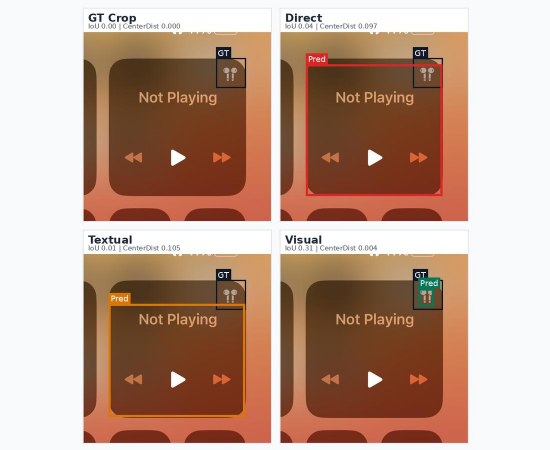}
        \qualcaptionfull{``View AirPods playback setting''}{D 0.04}{0.01}{0.31}
      \end{minipage}\hfill
      \begin{minipage}[t]{0.238\linewidth}
        \centering
        \includegraphics[width=\linewidth]{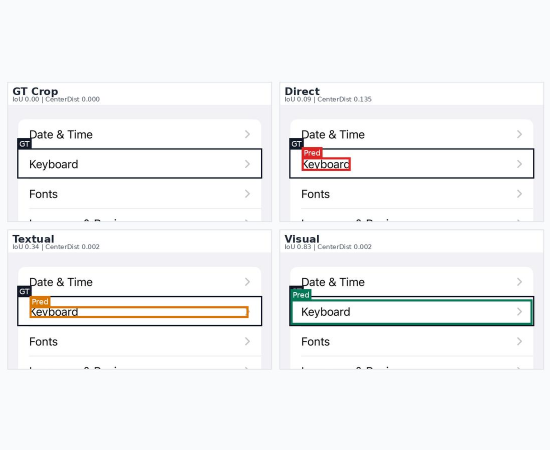}
        \qualcaptionfull{``check keyboard settings''}{D 0.09}{0.34}{0.83}
      \end{minipage}
    }}

  \vspace{5pt}

  \noindent\colorbox{VSqualCountHead}{%
    \parbox{\dimexpr\textwidth-2\fboxsep\relax}{%
      \normalsize\textbf{Counting Cases}
      \hfill {\color{VSqualMuted}dynamic visual priors externalize counting state}
    }}
  \vspace{2pt}

  \noindent\colorbox{VSqualCountBg}{%
    \parbox{\dimexpr\textwidth-2\fboxsep\relax}{%
      \begin{minipage}[t]{0.238\linewidth}
        \centering
        \includegraphics[width=\linewidth]{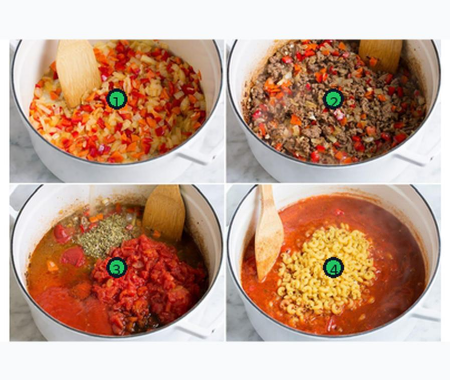}
        \qualcaptionfull{``How many pots?''}{GT 4; D 1}{1}{4}
      \end{minipage}\hfill
      \begin{minipage}[t]{0.238\linewidth}
        \centering
        \includegraphics[width=\linewidth]{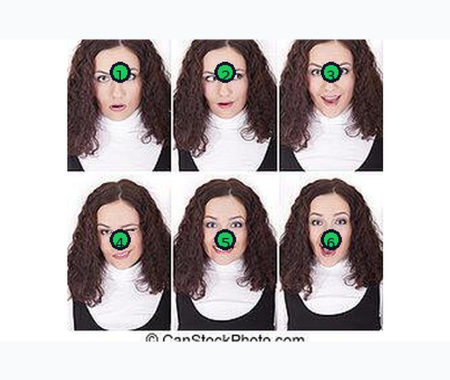}
        \qualcaptionfull{``How many women?''}{GT 6; D 1}{1}{6}
      \end{minipage}\hfill
      \begin{minipage}[t]{0.238\linewidth}
        \centering
        \includegraphics[width=\linewidth]{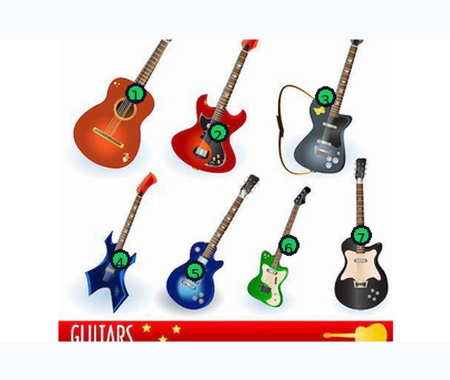}
        \qualcaptionfull{``How many guitars?''}{GT 7; D 8}{7}{7}
      \end{minipage}\hfill
      \begin{minipage}[t]{0.238\linewidth}
        \centering
        \includegraphics[width=\linewidth]{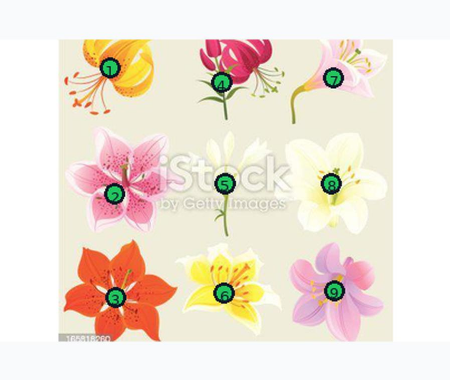}
        \qualcaptionfull{``How many flowers?''}{GT 9; D 9}{14}{9}
      \end{minipage}
    }}

  \caption{\textbf{Qualitative examples of visual skills.}
  Top: GUI grounding examples where visual priors improve hitbox localization.
  Bottom: dense-counting examples where dynamic visual traces improve count prediction.
  D, T, and V denote direct prompting, text-only skill prompting, and visual-skill prompting, respectively; GUI numbers are IoU, and counting numbers are predicted counts.}
  \label{fig:qualitative-visual-skill-cases}
\end{figure*}

The qualitative cases in \cref{fig:qualitative-visual-skill-cases} further illustrate why this mechanism is different from merely asking the model to be careful. In the GUI examples, the visual prior changes the target granularity from a vague semantic label to a visible hitbox convention, improving localization even when the textual instruction is short. In the counting examples, the rendered anchors expose which instances have already been selected, so the model can audit omissions, duplicate visits, and semantic granularity before returning the final count.

Together, the two controlled tasks isolate complementary failure modes of text-only skill reuse. Static priors recover reusable spatial conventions that are difficult to describe precisely, while dynamic priors maintain a visible intermediate state when the task requires repeated inspection. We therefore use these two benchmarks as canonical evidence for the primitive visual mechanisms, and use the broader qualitative gallery to show how the same mechanisms transfer to richer visual-skill forms.

\section{Discussion}

\subsection{\NAME vs.\ Image-based Few-shot Prompting}

\NAME differs from image-based few-shot prompting in both purpose and operational granularity. Few-shot examples are \emph{instance-level}: they provide input--output pairs from a related distribution and encourage local pattern imitation within that specific context. Visual priors are \emph{protocol-level}: they encode reusable conventions, such as target granularity, spatial boundaries, layout prototypes, and inspection procedures, without ever containing target answers for any given instance.

Thus, few-shot prompting acts as a temporary context cache, whereas \NAME provides persistent and reusable skill artifacts that can be retrieved, versioned, composed, and audited across diverse task instances. This distinction is particularly important for long-term agent capability accumulation, where reusable modules should be clearly decoupled from any single prompt instance.

\subsection{Will Stronger Models Eliminate the Need for Visual Protocols?}

Stronger models may reduce some failures, but they do not remove the need for modality-matched skill representations.
Visual skills are not merely a remedy for current model limitations; they preserve reusable spatial knowledge in a form that text alone inherently struggles to faithfully express.
When task rules depend on dense geometry, visual boundaries, or continuous perceptual tracking, encoding them purely as prose can introduce ambiguity and compression loss.

What stronger models may change is the \emph{form} of visual skills: static diagrams may evolve into videos, interactive annotations, executable visual programs, or learned visual memory modules. The core principle remains that reusable task knowledge with inherent visual structure should be preserved in a visual or multimodal form rather than discarded.

\subsection{Design Principles for Effective Visual Priors}

Not every image serves as an effective visual prior.
Our experiments suggest three design principles:

\begin{enumerate}[leftmargin=1.8em]
  \item \textbf{Abstract, not instance-specific.}
        A visual prior should distill spatial protocols shared across samples, rather than copy a test image or encode instance-specific answers.

  \item \textbf{Genuinely visual.}
        A visual prior should contain shapes, positions, boundaries, layouts, or spatial procedures that are difficult to express linearly.
        Screenshots of textual instructions usually provide little benefit over text itself.

  \item \textbf{Complementary to text.}
        Information that is already clear in language should remain in the textual logic.
        Visual priors should carry the spatial structure that text alone struggles to express.
\end{enumerate}

\subsection{Boundaries: When \emph{Not} to Use Visual Skills}

Visual skills are most useful when the task bottleneck is spatial or perceptual.
They may be unnecessary, or even distracting, in settings where knowledge is already well represented in language.

\begin{itemize}[leftmargin=1.8em]
  \item \textbf{Purely symbolic tasks}, such as algebraic computation, SQL generation, or code synthesis, where reusable knowledge is naturally discrete, procedural, and linguistic.

  \item \textbf{Unstructured open-ended perception}, such as free-form VQA over natural scenes, where imposing a rigid spatial schema may constrain the model's native visual reasoning.
\end{itemize}

\noindent Overall, visual skills should be invoked when the reusable task knowledge contains inherent spatial structure that is difficult to accurately preserve through text alone without loss.

\section{Conclusion}
We identified the \textit{textual bottleneck} in current agent skill paradigms, where reusable experience is stored mainly as text despite the inherently visual and spatial nature of many agent tasks. To address this fundamental mismatch, we proposed \textbf{\NAME}, a reusable multimodal skill entity that combines declarative textual logic, visual priors or references, and a multimodal binding protocol. We further introduced \textbf{\SYSTEM}, a proof-of-concept authoring pipeline that automatically diagnoses visual bottlenecks, generates textual and visual skill components, and packages them into reusable \NAME artifacts. Experiments on GUI grounding and dense counting consistently show that visual skills outperform text-only skills, especially when success requires spatial conventions, localized visual evidence, and grounded intermediate tracking. Beyond the two evaluated primitive prior families, we also define interleaved visual skills for cases where ordered reasoning must remain adjacent to the visual evidence that grounds each step. These findings suggest that reusable agent skills should go beyond text: textual logic should provide high-level procedural control, while visual priors and references should preserve the spatial structure needed for grounding, inspection, verification, and source-grounded explanation. Together, \NAME and \SYSTEM point toward more reusable, composable, and inspectable skill representations for the next generation of multimodal agents.

{\small
\bibliographystyle{plainnat}
\bibliography{ref}
}
%%%%%%%%%%%%%%%%%%%%%%%%%%%%%%%%%%%%%%%%%%%%%%%%%%%%%%%%%%%%

\appendix

% Appendix snippet for Visual Skill protocol details.
% Expected packages in the main paper: booktabs, tabularx, array, amsmath.

\section{Technical appendices and supplementary material}
\label{app:protocol-details}

\subsection{Complete Prompt Templates}
\label{app:prompt-templates}

All settings share the same task image, question, model, decoding parameters, and output parser. The only intervention is whether the model receives no reusable skill, text-only reusable logic, or the full visual-skill artifact.

\begin{promptbox}[GUI Grounding: Direct Prompting]
In this GUI screenshot, locate the point to click for the instruction:\\
\pvar{"\{instruction\}"}

\promptgap
Return only JSON:\\
\pcode{\{"point\_2d":[x,y], "bbox\_2d":[x1,y1,x2,y2]\}}

\promptgap
All coordinates are integers in the 0--1000 coordinate system relative to
the task screenshot. The point should be the click point, and the bbox
should cover the target clickable region.
\end{promptbox}

\begin{promptbox}[GUI Grounding: Textual Skill]
You are applying the GUI grounding textual skill.

\promptgap
Declarative textual logic:\\
\pvar{\{gui\_textual\_skill\_rules\}}

\promptgap
Task screenshot:\\
\pph{[INSERT GUI SCREENSHOT HERE]}

\promptgap
Instruction: \pvar{"\{instruction\}"}

\promptgap
Return only valid JSON:\\
\pcode{\{"point\_2d":[x,y], "bbox\_2d":[x1,y1,x2,y2]\}}

\promptgap
All coordinates are integers in the 0--1000 coordinate system relative only
to the task screenshot.
\end{promptbox}

\begin{promptbox}[GUI Grounding: Visual Skill]
You are applying the GUI grounding visual skill.

\promptgap
Visual skill artifact:\\
\pvar{\{gui\_visual\_skill\_artifact\}}

\promptgap
\pcode{<visual\_prior>}\\
\pph{[INSERT VISUAL PRIOR IMAGE HERE]}\\
\pcode{</visual\_prior>}

\promptgap
\pcode{<task\_screenshot>}\\
\pph{[INSERT GUI SCREENSHOT HERE]}\\
\pcode{</task\_screenshot>}

\promptgap
Instruction: \pvar{"\{instruction\}"}

\promptgap
Respond only with valid JSON:\\
\pcode{\{"point\_2d":[x,y], "bbox\_2d":[x1,y1,x2,y2]\}}

\promptgap
All coordinates must refer only to the task screenshot. The visual prior is
a reusable protocol reference and has no answer coordinates.
\end{promptbox}

\begin{promptbox}[CountBenchQA: Direct Prompting]
Question: \pvar{\{question\}}

\promptgap
Count the target objects in the image.\\
Return one line of valid JSON only:\\
\pcode{\{"total\_count": N\}}
\end{promptbox}

\begin{promptbox}[CountBenchQA: Textual Skill]
Use the counting textual skill.

\promptgap
Declarative textual logic:\\
\pvar{\{counting\_textual\_skill\_rules\}}

\promptgap
Task image:\\
\pph{[INSERT COUNTING IMAGE HERE]}

\promptgap
Return one line of valid JSON only:\\
\pcode{\{"total\_count": N\}}

\promptgap
Question: \pvar{\{question\}}
\end{promptbox}

\begin{promptbox}[CountBenchQA: Visual Skill with Dynamic Prior]
Use the counting visual skill with dynamic prior feedback.

\promptgap
Visual skill artifact:\\
\pvar{\{counting\_visual\_skill\_artifact\}}

\promptgap
Current task image:\\
\pph{[INSERT CURRENT IMAGE, INCLUDING ANY DYNAMIC ANCHORS]}

\promptgap
Return one line of valid JSON only:\\
\pcode{\{"points\_2d":[[x,y], ...], "total\_count": N\}}

\promptgap
Question: \pvar{\{question\}}
\end{promptbox}

\noindent
\begin{minipage}{\linewidth}
\small\itshape\color{pphcolor}%
For the dynamic-prior setting, the predicted points are rendered back onto the task image as numbered visual anchors. The next request receives only the newly rendered task image and the same fixed prompt, so the marked image becomes external visual working memory. The loop stops when the model returns no new points or when a conservative maximum-round limit is reached.
\end{minipage}

\begin{promptbox}[CountQA: Direct Prompting]
Question: \pvar{\{question\}}

\promptgap
Count the target objects in the image.\\
Return one line of valid JSON only:\\
\pcode{\{"total\_count": N\}}
\end{promptbox}

\begin{promptbox}[CountQA: Textual Skill]
Use the counting textual skill.

\promptgap
Declarative textual logic:\\
\pvar{\{counting\_textual\_skill\_rules\}}

\promptgap
Task image:\\
\pph{[INSERT COUNTING IMAGE HERE]}

\promptgap
Return one line of valid JSON only:\\
\pcode{\{"total\_count": N\}}

\promptgap
Question: \pvar{\{question\}}
\end{promptbox}

\begin{promptbox}[CountQA: Visual Skill with Dynamic Prior]
Use the counting visual skill with dynamic prior feedback.

\promptgap
Visual skill artifact:\\
\pvar{\{counting\_visual\_skill\_artifact\}}

\promptgap
Current task image:\\
\pph{[INSERT CURRENT IMAGE, INCLUDING ANY DYNAMIC ANCHORS]}

\promptgap
Question: \pvar{\{question\}}

\promptgap
Return one line of valid JSON only:\\
\pcode{\{"points\_2d":[[x,y], ...], "total\_count": N\}}
\end{promptbox}

\subsection{Metric Definitions}
\label{app:metric-definitions}

\paragraph{GUI grounding.}
Let the ground-truth box be $B^\star=[x_1^\star,y_1^\star,x_2^\star,y_2^\star]$, the predicted box be $\hat{B}$, and the predicted click point be $\hat{p}=(\hat{x},\hat{y})$.

\begin{align}
\mathrm{PointInBox}(\hat{p}, B^\star)
&= \mathbb{1}\left[x_1^\star \le \hat{x} \le x_2^\star
\,\wedge\, y_1^\star \le \hat{y} \le y_2^\star\right], \\
\mathrm{IoU}(\hat{B}, B^\star)
&= \frac{\mathrm{area}(\hat{B}\cap B^\star)}
{\mathrm{area}(\hat{B}\cup B^\star)}, \\
\mathrm{CenterDist}(\hat{B}, B^\star)
&= \frac{\left\|c(\hat{B}) - c(B^\star)\right\|_2}
{\sqrt{W^2+H^2}},
\end{align}
where $c(\cdot)$ is the box center and $(W,H)$ is the image size. Higher Point-in-Box and IoU are better; lower Center Distance is better.

\paragraph{Counting.}
Let $y_i$ be the ground-truth count and $\hat{y}_i$ be the predicted count.

\begin{align}
\mathrm{ExactAcc}
&= \frac{1}{N}\sum_{i=1}^N \mathbb{1}[\hat{y}_i = y_i], \\
\mathrm{MAE}
&= \frac{1}{N}\sum_{i=1}^N |\hat{y}_i-y_i|, \\
\mathrm{Within1}
&= \frac{1}{N}\sum_{i=1}^N \mathbb{1}[|\hat{y}_i-y_i|\le 1].
\end{align}

For visual-skill counting, we additionally check internal consistency when point anchors are returned:
\begin{equation}
\mathrm{AnchorConsistency}
= \mathbb{1}\left[|\mathrm{points\_2d}|=\hat{y}\right].
\end{equation}
This diagnostic is not a replacement for count accuracy; it only verifies that the reported count agrees with the number of visual anchors emitted by the model.

\subsection{Failure Mode Analysis: The Limits of Visual Protocols}
\label{app:failure-mode-analysis}

\begin{figure}[H]
  \centering
  \scriptsize
  \setlength{\fboxsep}{5pt}
  \definecolor{VSredHead}{HTML}{FAD1D1}
  \definecolor{VSredBg}{HTML}{FFF3F3}
  \definecolor{VSorangeHead}{HTML}{FFE2B8}
  \definecolor{VSorangeBg}{HTML}{FFF8ED}
  \definecolor{VSquestion}{HTML}{FFF4C2}
  \definecolor{VSred}{HTML}{B42318}
  \definecolor{VSmuted}{HTML}{6B7280}

  \newcommand{\failcaptionapp}[3]{%
    \vspace{1pt}
    \colorbox{VSquestion}{\parbox{0.94\linewidth}{\centering\textit{#1}}}\\[-1pt]
    \parbox{0.96\linewidth}{\centering #2}\\[-1pt]
    \parbox{0.96\linewidth}{\centering{\color{VSred}#3}}
  }

  \noindent\colorbox{VSredHead}{%
    \parbox{\dimexpr\textwidth-2\fboxsep\relax}{%
      \normalsize\textbf{GUI Grounding Failure Cases}
      \hfill {\color{VSmuted}when visual priors over-specialize target granularity}
    }}
  \vspace{2pt}

  \noindent\colorbox{VSredBg}{%
    \parbox{\dimexpr\textwidth-2\fboxsep\relax}{%
      \begin{minipage}[t]{0.238\linewidth}
        \centering
        \includegraphics[width=\linewidth]{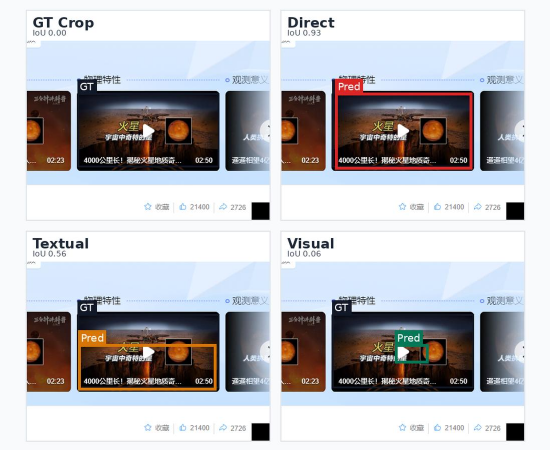}
        \failcaptionapp{``play the Mars video''}{D 0.93; T 0.56; V 0.06}{over-focuses on inner play glyph}
      \end{minipage}\hfill
      \begin{minipage}[t]{0.238\linewidth}
        \centering
        \includegraphics[width=\linewidth]{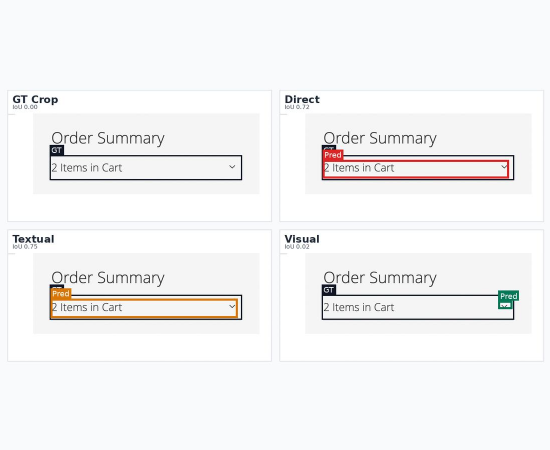}
        \failcaptionapp{``show more items in cart''}{D 0.72; T 0.75; V 0.02}{selects chevron, not full selector}
      \end{minipage}\hfill
      \begin{minipage}[t]{0.238\linewidth}
        \centering
        \includegraphics[width=\linewidth]{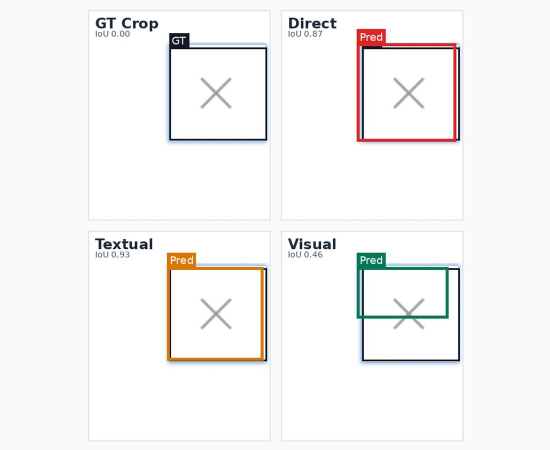}
        \failcaptionapp{``close the image window''}{D 0.87; T 0.93; V 0.46}{hitbox boundary is shifted upward}
      \end{minipage}\hfill
      \begin{minipage}[t]{0.238\linewidth}
        \centering
        \includegraphics[width=\linewidth]{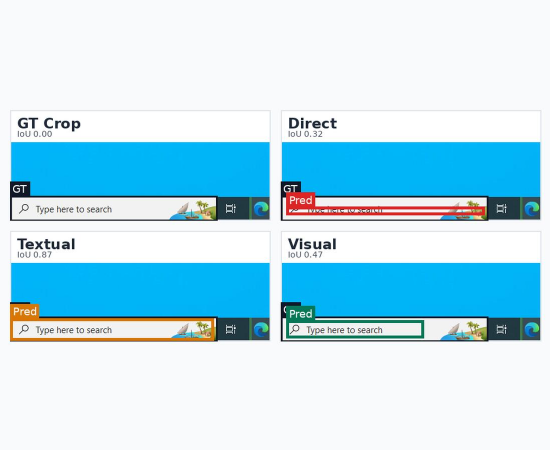}
        \failcaptionapp{``click the search bar''}{D 0.32; T 0.87; V 0.47}{underestimates field width}
      \end{minipage}
    }}

  \vspace{5pt}

  \noindent\colorbox{VSorangeHead}{%
    \parbox{\dimexpr\textwidth-2\fboxsep\relax}{%
      \normalsize\textbf{Counting Failure Cases}
      \hfill {\color{VSmuted}when visual trajectories expose semantic or granularity ambiguity}
    }}
  \vspace{2pt}

  \noindent\colorbox{VSorangeBg}{%
    \parbox{\dimexpr\textwidth-2\fboxsep\relax}{%
      \begin{minipage}[t]{0.238\linewidth}
        \centering
        \includegraphics[width=\linewidth]{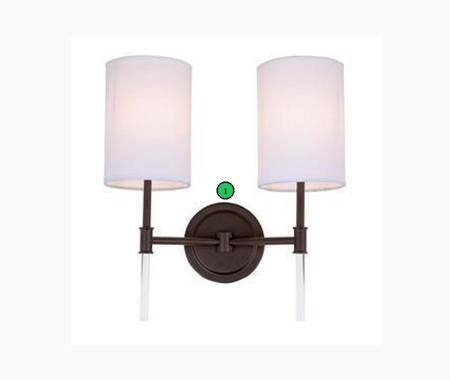}
        \failcaptionapp{``How many sconces?''}{GT 2; D 1; T 1; V 1}{composite fixture treated as one}
      \end{minipage}\hfill
      \begin{minipage}[t]{0.238\linewidth}
        \centering
        \includegraphics[width=\linewidth]{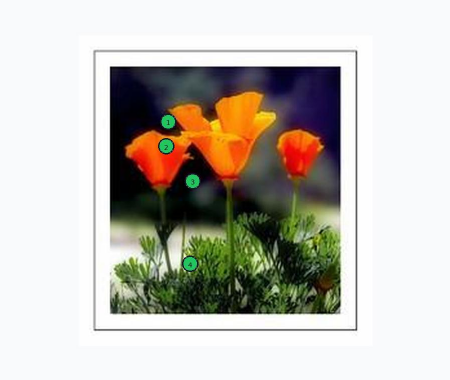}
        \failcaptionapp{``How many flowers?''}{GT 3; D 4; T 4; V 4}{foliage/partial region counted}
      \end{minipage}\hfill
      \begin{minipage}[t]{0.238\linewidth}
        \centering
        \includegraphics[width=\linewidth]{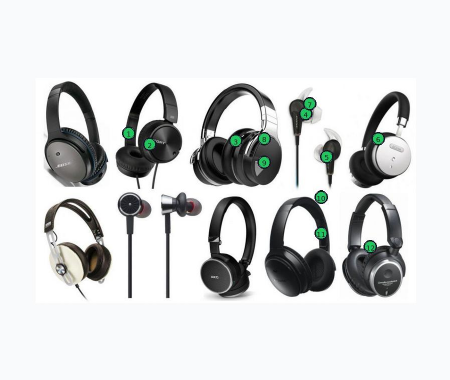}
        \failcaptionapp{``How many headphone sets?''}{GT 10; D 10; T 10; V 12}{set decomposed into subparts}
      \end{minipage}\hfill
      \begin{minipage}[t]{0.238\linewidth}
        \centering
        \includegraphics[width=\linewidth]{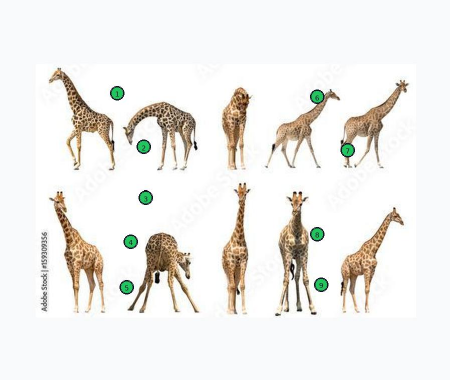}
        \failcaptionapp{``How many giraffes?''}{GT 10; D 10; T 10; V 9}{misses a low-contrast instance}
      \end{minipage}
    }}

  \caption{\textbf{Representative failure cases.}
  Visual Skills can fail when the intended spatial convention conflicts with semantic granularity: GUI priors may over-focus on the smallest glyph-like component, while counting trajectories may expose ambiguity between composite objects, subparts, and low-salience instances.}
  \label{fig:visual-skill-failure-cases}
\end{figure}

While Visual Skills substantially mitigate the textual bottleneck, they are not immune to errors. As illustrated in Figure~\ref{fig:visual-skill-failure-cases}, enforcing explicit visual protocols occasionally introduces new classes of failure, primarily stemming from a tension between structural spatial priors and fine-grained semantic intent.

\paragraph{GUI Grounding: Over-specialization of Target Granularity.}
In static-prior settings, the visual protocol can over-enforce a specific spatial convention, such as bounding the minimal clickable icon. This strong structural bias may override the semantic scope of instructions such as \enquote{play the Mars video} or \enquote{show more items in cart}, causing the model to focus on a small glyph rather than the full functional container.

\paragraph{Dense Counting: Semantic and Granularity Ambiguity.}
Dynamic priors externalize spatial memory, but they also force the model to commit to what counts as one discrete object. This exposes errors such as treating a composite fixture as one object, decomposing a headset into subparts, counting background foliage, or missing low-contrast instances. These failures suggest that future visual-skill systems should better arbitrate between textual semantic scope and rigid spatial schemas.

\subsection{Complete Visual Skill Artifact Specifications}
\label{sec:visual-skill-specifications}

To provide concrete examples of the multimodal assets evaluated in Section 5, we present the complete Visual Skill specifications generated by the \SYSTEM pipeline. These artifacts explicitly demonstrate our core design principle: delegating abstract reasoning and boundary conditions to the textual modality, while isolating spatial conventions and tracking mechanisms within the visual modality. 

Table~\ref{tab:gui-visual-skill-case-spec} details the static Visual Skill utilized for the GUI grounding task. The textual logic outlines the target filtering procedure, while the static visual prior serves as a global dictionary, calibrating the model's understanding of implicit hitboxes and structural UI boundaries. 

Table~\ref{tab:count-dynamic-skill-spec} presents the dynamic Visual Skill deployed for the dense object counting task. In this specification, the textual rules govern semantic inclusion and exclusion criteria (e.g., ignoring reflections or sub-parts), while the dynamic prior enforces an \textit{in-situ} spatial anchoring protocol to maintain a reliable visual working memory during iterative enumeration.

%% ===== Table 1: Icon Boundary Protocol Spec =====
\begin{table*}[t!]
\centering
\caption{\textbf{Visual Skill specification for GUI grounding.} The icon grounding skill uses a reusable visual before binding textual rules to spatial click protocols.}
\label{tab:gui-visual-skill-case-spec}
\vspace{0.8em}

\begin{tcolorbox}[
  enhanced,
  colback=cloudwhite,
  colframe=babyblueborder,
  boxrule=1.2pt,
  arc=14pt,
  outer arc=14pt,
  left=4pt, right=4pt, top=5pt, bottom=5pt,
  boxsep=3pt,
  toptitle=10pt, bottomtitle=10pt,
  title={\Large\sffamily\bfseries\textcolor{softnavytitle}{\faIcon{crosshairs}\hspace{0.6em}Icon Boundary Protocol for GUI Grounding}},
  coltitle=softnavytitle,
  colbacktitle=babyblue,
  fonttitle=\large\sffamily\bfseries,
  fuzzy shadow={0mm}{-2mm}{0mm}{0.4mm}{babyblueborder!40},
  borderline={0.4pt}{1pt}{babyblueborder!30},
]
\small\setstretch{1.28}

\begin{tabular}{@{}%
  >{\centering\arraybackslash}m{0.16\linewidth}%
  @{\hspace{0.03\linewidth}}%
  >{\RaggedRight\arraybackslash}m{0.77\linewidth}%
  @{}}
\\[-0.3em]

%% --- Row 1: Description ---
\sectionlabel{Description}
&
\begin{minipage}[c]{\linewidth}
\textcolor{textmain}{Locate small GUI icon controls by selecting the complete clickable hitbox rather than only the visible glyph. The skill is designed for toolbar icons, window controls, row actions, and icon-only buttons.}
\end{minipage}
\\[0.9em]

\arrayrulecolor{babyblueborder!40}\hline\\[0.45em]

%% --- Row 2: Visual Prior ---
\sectionlabel{Visual Prior}
&
\begin{minipage}[c]{\linewidth}
\vspace{0.2em}
\centering
\begin{tcolorbox}[
  enhanced,
  colback=snowgray,
  colframe=babyblueborder!50,
  boxrule=0.8pt,
  arc=10pt,
  outer arc=10pt,
  boxsep=4pt,
  left=5pt, right=5pt, top=5pt, bottom=5pt,
  width=0.56\linewidth,
  fuzzy shadow={0mm}{-1mm}{0mm}{0.3mm}{babyblueborder!30},
]
\centering
\includegraphics[width=\linewidth]{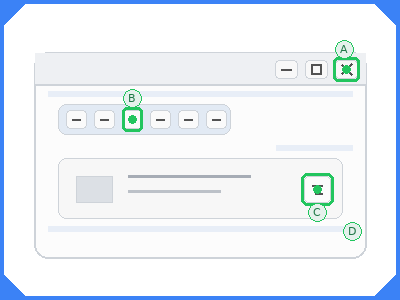}
\end{tcolorbox}
\vspace{0.35em}

\RaggedRight
\textcolor{textfaint}{The prior is abstract and benchmark-safe. \textcolor{mintdark}{\textbf{Green}} marks the selected hitbox center; neighboring candidates and parent containers are shown only as structural context, not as coordinates to copy.}
\end{minipage}
\\[0.9em]

\arrayrulecolor{babyblueborder!40}\hline\\[0.45em]

%% --- Row 3: Textual Logic ---
\sectionlabel{Textual Logic}
&
\begin{minipage}[c]{\linewidth}
\vspace{0.1em}
\begin{enumerate}[leftmargin=*, itemsep=3pt, label={\textcolor{babybluedeep}{\sffamily\bfseries\arabic*.}}]
\item Match the instruction to one icon-like functional target.
\item Estimate the padded clickable hitbox around the glyph.
\item Resolve adjacent icons by local cluster context.
\item If nested in a row/card, choose the smallest child action, not the parent container.
\item Return the hitbox center and a tight \texttt{bbox\_2d} in the task screenshot coordinate space.
\end{enumerate}
\vspace{0.1em}
\end{minipage}
\\[0.3em]

\end{tabular}
\end{tcolorbox}
\end{table*}

\begin{table*}[t!]
\centering
\caption{\textbf{Dynamic Visual Skill for CountBenchQA.} The visual skill converts counting into point-anchored enumeration.}
\label{tab:count-dynamic-skill-spec}
\vspace{0.8em}

\begin{tcolorbox}[
  enhanced,
  colback=cloudwhite,
  colframe=mintborder,
  boxrule=1.2pt,
  arc=14pt,
  outer arc=14pt,
  left=4pt, right=4pt, top=5pt, bottom=5pt,
  boxsep=3pt,
  toptitle=10pt, bottomtitle=10pt,
  title={\Large\sffamily\bfseries\textcolor{mintdark}{\faIcon{map-marker-alt}\hspace{0.6em}Point-Anchored Enumeration for Visual Counting}},
  coltitle=mintdark,
  colbacktitle=mintcream,
  fonttitle=\large\sffamily\bfseries,
  fuzzy shadow={0mm}{-2mm}{0mm}{0.4mm}{mintborder!40},
  borderline={0.4pt}{1pt}{mintborder!30},
]
\small\setstretch{1.28}

\begin{tabular}{@{}%
  >{\centering\arraybackslash}m{0.16\linewidth}%
  @{\hspace{0.03\linewidth}}%
  >{\RaggedRight\arraybackslash}m{0.77\linewidth}%
  @{}}
\\[-0.3em]

%% --- Row 1: Description ---
\sectionlabel{Description}
&
\begin{minipage}[c]{\linewidth}
\textcolor{textmain}{Count dense or ambiguous objects by producing one spatial anchor per valid target instance. The anchors are rendered back onto the image as a dynamic visual prior, so later reasoning can see which instances have already been counted.}
\end{minipage}
\\[0.9em]

\arrayrulecolor{mintborder!40}\hline\\[0.45em]

%% --- Row 2: Textual Logic ---
\sectionlabel{Textual Logic}
&
\begin{minipage}[c]{\linewidth}
\vspace{0.1em}
\begin{enumerate}[leftmargin=*, itemsep=3pt, label={\textcolor{mintdeep}{\sffamily\bfseries\arabic*.}}]
\item Identify the target noun and its semantic granularity.
\item Place exactly one point at the center of each physical instance.
\item Ignore legends, labels, embedded media, reflections, fragments, and sub-parts unless explicitly requested.
\item Scan exhaustively; low-contrast or partially occluded valid instances still receive anchors.
\item Report \texttt{points\_2d} and \texttt{total\_count}; the count is the length of the anchored instance list.
\end{enumerate}
\vspace{0.1em}
\end{minipage}
\\[0.9em]

\arrayrulecolor{mintborder!40}\hline\\[0.45em]

%% --- Row 3: Dynamic Prior ---
\sectionlabel{Dynamic Prior}
&
\begin{minipage}[c]{\linewidth}
\textcolor{textmain}{The prior is not a fixed image template. It is generated \textbf{during inference} by overlaying numbered \textcolor{mintdark}{\textbf{green anchors}} on the task image. This makes the model's counting process visual, auditable, and reusable across object categories.}
\end{minipage}
\\[0.3em]

\end{tabular}
\end{tcolorbox}
\end{table*}

\subsection{Decision Boundaries for Modality Selection}
\label{sec:prior-decision-boundaries}

To systematically operationalize the \textbf{\NAME} paradigm, we establish a rigorous taxonomy for modality selection, detailed in Figure~\ref{fig:prior-decision}. The core premise of this decision matrix is that the optimal skill representation must be dictated by the underlying \textit{cognitive bottleneck} of the task, rather than its domain or dataset label. 

Rather than arbitrarily injecting visual elements into all agent workflows, Figure~\ref{fig:prior-decision} outlines an explicit routing logic. It delineates the exact boundaries for when to rely on text-only rules, when to invoke static or dynamic visual priors, when to use interleaved step-to-evidence bindings, and when a combined approach is necessary. This taxonomy ensures that multimodal assets are deployed purposefully---decisively breaking the textual bottleneck in spatially intensive environments, while preventing extraneous cognitive noise in strictly logical tasks.

\begin{figure*}[t]
\centering

\begin{choicebox}[title={1 \enspace Text-only rules},
                   colframe=blue!50!black, colbacktitle=blue!10]
\begin{description}[leftmargin=7em, style=nextline, font=\normalfont\itshape]
  \item[Use when] The relevant constraint is discrete, symbolic, or more precise in language than in an image.
  \item[Avoid when] The model must infer a visual boundary, target granularity, or remembered spatial state.
  \item[Examples] Output schema, confidence gates, category exclusion rules.
\end{description}
\end{choicebox}

\vspace{4pt}

\begin{choicebox}[title={2 \enspace Static visual prior},
                   colframe=teal!60!black, colbacktitle=teal!10]
\begin{description}[leftmargin=7em, style=nextline, font=\normalfont\itshape]
  \item[Use when] The bottleneck is a reusable spatial convention or protocol ambiguity.
  \item[Avoid when] The task requires tracking which instances have already been processed.
  \item[Examples] GUI hitbox vs.\ glyph, nested child control vs.\ parent container.
\end{description}
\end{choicebox}

\vspace{4pt}

\begin{choicebox}[title={3 \enspace Dynamic visual prior},
                   colframe=orange!70!black, colbacktitle=orange!10]
\begin{description}[leftmargin=7em, style=nextline, font=\normalfont\itshape]
  \item[Use when] The bottleneck is perceptual tracking or visual working memory over repeated steps.
  \item[Avoid when] A single static rule already resolves the ambiguity.
  \item[Examples] Dense counting with incremental anchors, route tracing with visited-state overlays.
\end{description}
\end{choicebox}

\vspace{4pt}

\begin{choicebox}[title={4 \enspace Interleaved visual skill},
                   colframe=violet!70!black, colbacktitle=violet!10]
\begin{description}[leftmargin=7em, style=nextline, font=\normalfont\itshape]
  \item[Use when] Ordered reasoning steps must remain adjacent to the source frame, screenshot, page, or crop that grounds each claim.
  \item[Avoid when] A single reusable prior or a pure text procedure is sufficient.
  \item[Examples] Visual tutorials, documentation workflows, PDF/slide walkthroughs, evidence-bound explanations.
\end{description}
\end{choicebox}

\vspace{4pt}

\begin{choicebox}[title={5 \enspace Combined use},
                   colframe=purple!60!black, colbacktitle=purple!10]
\begin{description}[leftmargin=7em, style=nextline, font=\normalfont\itshape]
  \item[Use when] A static convention defines what should be marked, and dynamic feedback stores which parts are already processed.
  \item[Avoid when] The added visual state distracts from a simple one-shot recognition task.
  \item[Examples] Count one object once, then render anchors back onto the image to prevent recounting.
\end{description}
\end{choicebox}
\caption{\textbf{When to use text-only rules, static priors, dynamic priors, or interleaved visual skills.} The boundary is determined by the type of bottleneck, not by the dataset name.}
\label{fig:prior-decision}

\end{figure*}

\FloatBarrier
\subsection{Visual Skill Example Gallery}
\label{app:example-gallery}

The examples below illustrate the three visual-skill forms discussed in the main text. The gallery is organized by capability: clarify spatial conventions, externalize runtime state, and bind reasoning steps to visual evidence.

\begin{figure*}[t]
\centering
\scriptsize
\setlength{\fboxsep}{2pt}
\begin{minipage}[t]{0.32\linewidth}
\begin{casecard}{babyblueborder}
\casebadge{caseStatic}{STATIC}\quad\textbf{Button click target}\\[3pt]
\includegraphics[width=\linewidth]{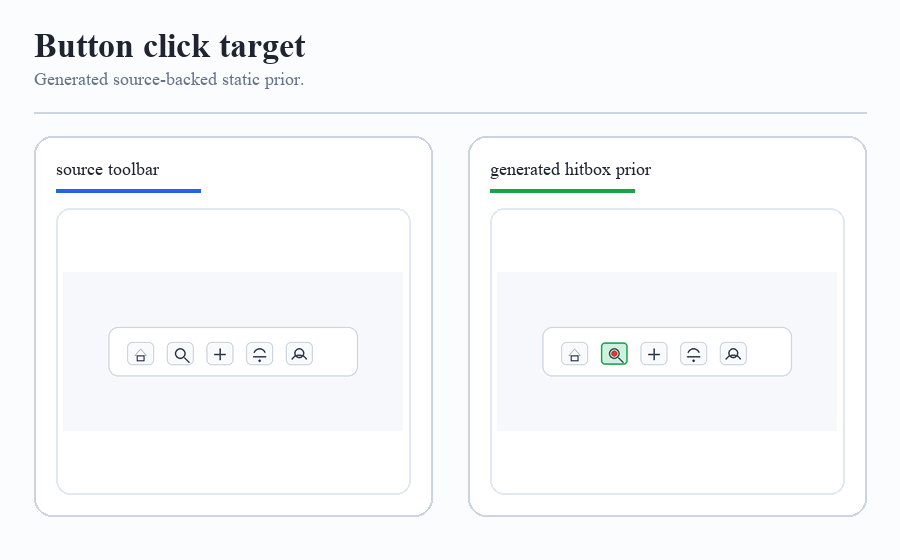}\\[-1pt]
\textcolor{caseMuted}{\textbf{Bottleneck.}} Small glyph ink is not identical to the clickable hitbox.\\
\textcolor{caseMuted}{\textbf{Visual support.}} A sparse overlay marks the button envelope and click center.
\end{casecard}
\end{minipage}\hfill
\begin{minipage}[t]{0.32\linewidth}
\begin{casecard}{babyblueborder}
\casebadge{caseStatic}{STATIC}\quad\textbf{Table cell intersection}\\[3pt]
\includegraphics[width=\linewidth]{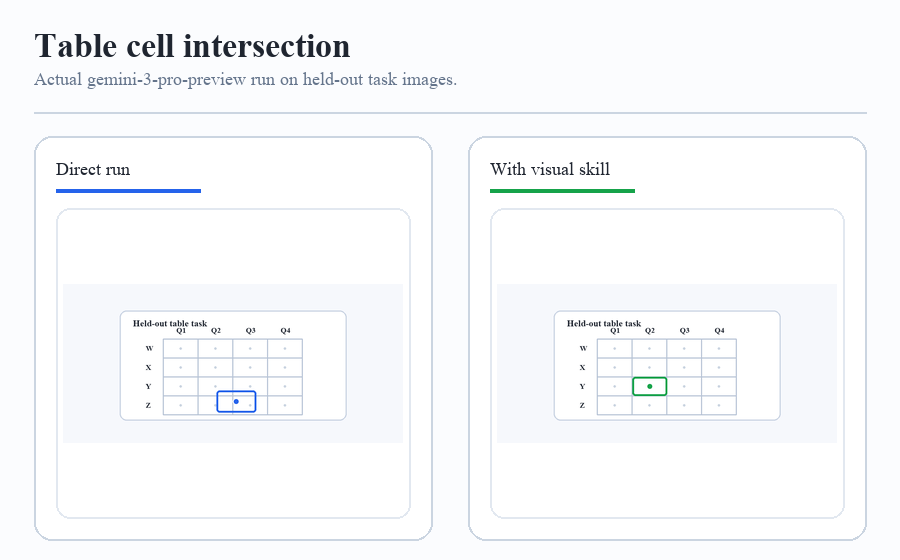}\\[-1pt]
\textcolor{caseMuted}{\textbf{Bottleneck.}} Row and column labels must be converted into an exact cell region.\\
\textcolor{caseMuted}{\textbf{Visual support.}} Horizontal and vertical bands make the intersection visible.
\end{casecard}
\end{minipage}\hfill
\begin{minipage}[t]{0.32\linewidth}
\begin{casecard}{babyblueborder}
\casebadge{caseStatic}{STATIC}\quad\textbf{Bar chart projection}\\[3pt]
\includegraphics[width=\linewidth]{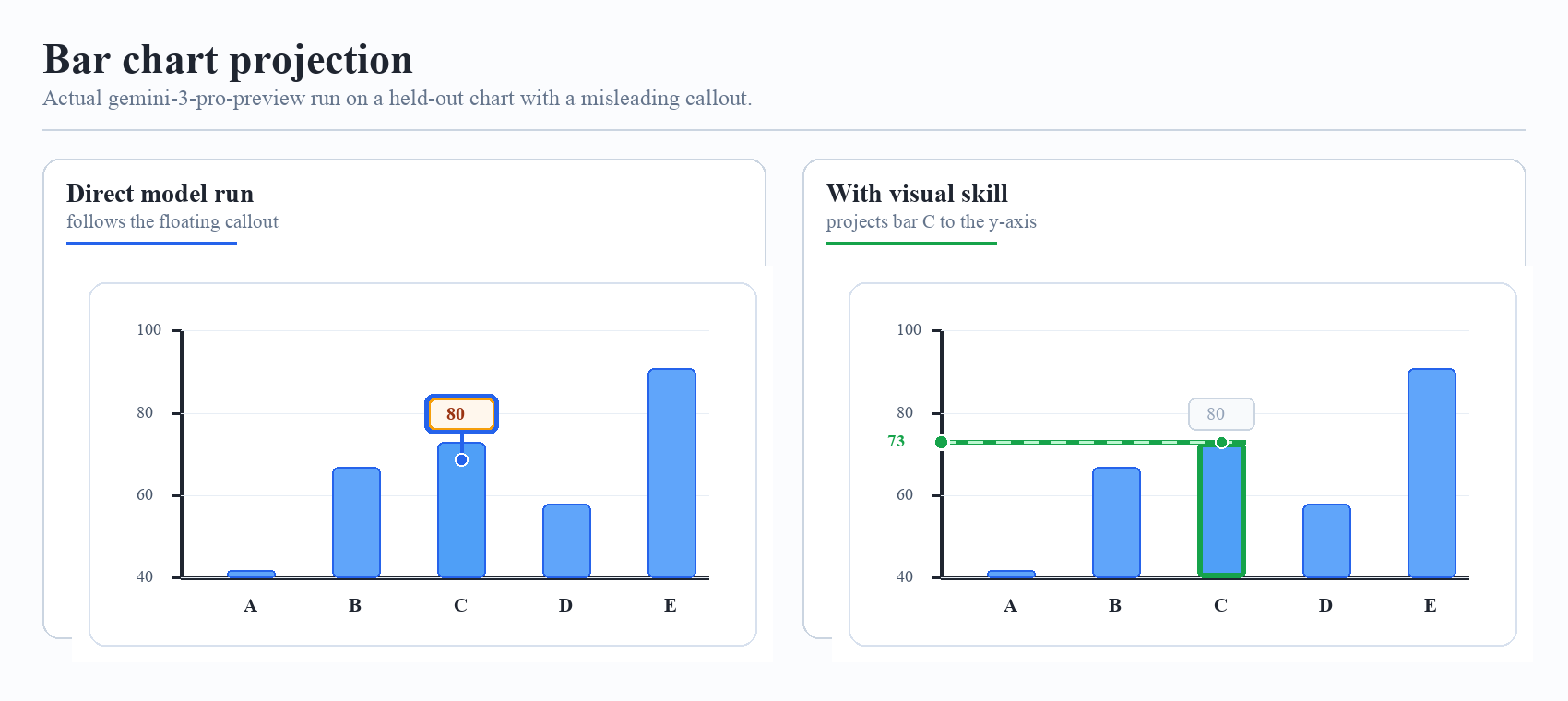}\\[-1pt]
\textcolor{caseMuted}{\textbf{Bottleneck.}} The value should come from geometric axis projection, not nearby labels.\\
\textcolor{caseMuted}{\textbf{Visual support.}} A selected bar, horizontal guide, and axis point encode the readout protocol.
\end{casecard}
\end{minipage}
\caption{\textbf{Static visual skill examples.} These examples use source-backed overlays to clarify reusable spatial conventions.}
\label{fig:appendix-static-cases}
\end{figure*}

\begin{figure*}[t]
\centering
\scriptsize
\begin{casecard}{peachborder}
\casebadge{caseApplied}{DYNAMIC/APPLIED}\quad\textbf{Presentation design critique-and-redraw}\\[3pt]
\includegraphics[width=\linewidth]{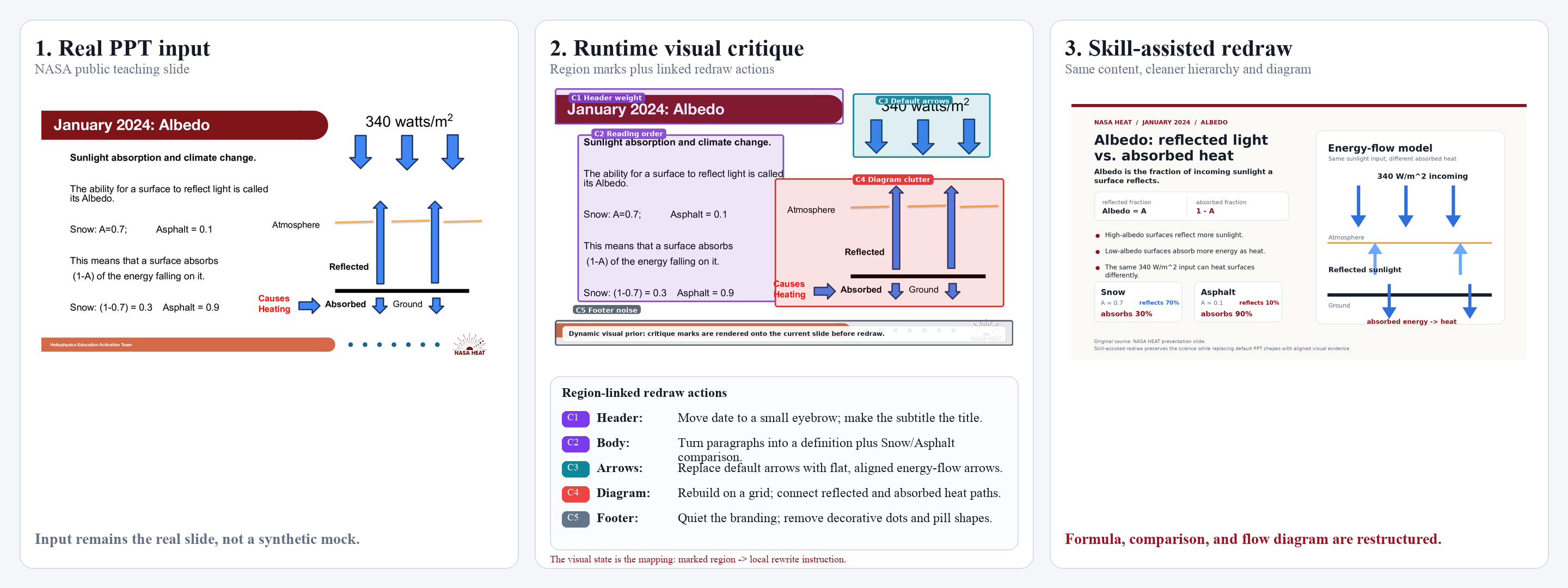}\\[-1pt]
\textcolor{caseMuted}{\textbf{Bottleneck.}} Text-only design advice often remains global and underspecified. The visual skill renders critique regions directly on the draft slide, binding each problem area to a local redraw action.
\end{casecard}

\vspace{4pt}
\begin{minipage}[t]{0.49\linewidth}
\begin{casecard}{mintborder}
\casebadge{caseDynamic}{DYNAMIC}\quad\textbf{Dense counting}\\[3pt]
\includegraphics[width=\linewidth]{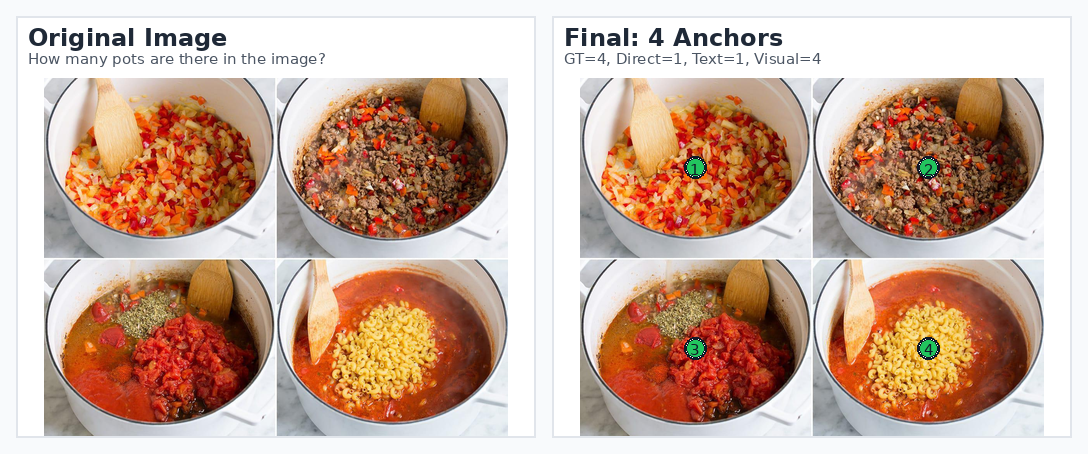}\\[-1pt]
\textcolor{caseMuted}{Anchors are rendered back onto counted instances so later calls can inspect visible progress rather than hidden memory.}
\end{casecard}
\end{minipage}\hfill
\begin{minipage}[t]{0.49\linewidth}
\begin{casecard}{mintborder}
\casebadge{caseDynamic}{DYNAMIC}\quad\textbf{Line tracing}\\[3pt]
\includegraphics[width=\linewidth]{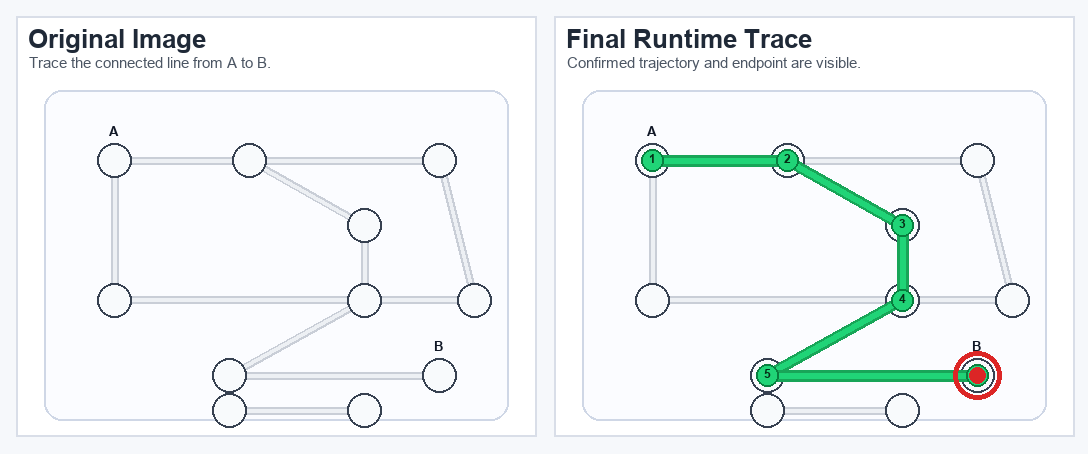}\\[-1pt]
\textcolor{caseMuted}{The current trajectory and endpoint are written back onto the graph so each step continues from visible path state.}
\end{casecard}
\end{minipage}

\vspace{2pt}
\begin{minipage}[t]{0.49\linewidth}
\begin{casecard}{mintborder}
\casebadge{caseDynamic}{DYNAMIC}\quad\textbf{Geometry auxiliary lines}\\[3pt]
\includegraphics[width=\linewidth]{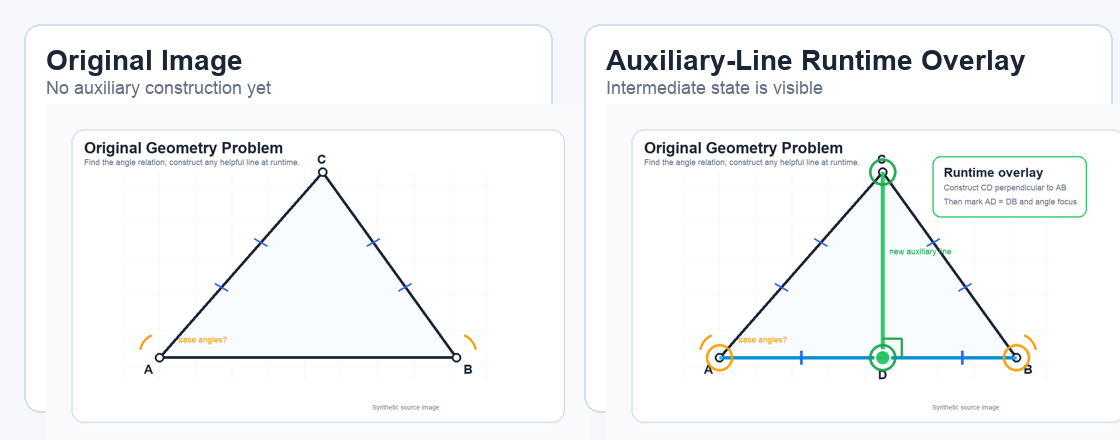}\\[-1pt]
\textcolor{caseMuted}{Auxiliary constructions, equalities, and angle focus are rendered on the current diagram as proof-state memory.}
\end{casecard}
\end{minipage}\hfill
\begin{minipage}[t]{0.49\linewidth}
\begin{casecard}{mintborder}
\casebadge{caseDynamic}{DYNAMIC}\quad\textbf{Odd-one-out visual search}\\[3pt]
\includegraphics[width=\linewidth]{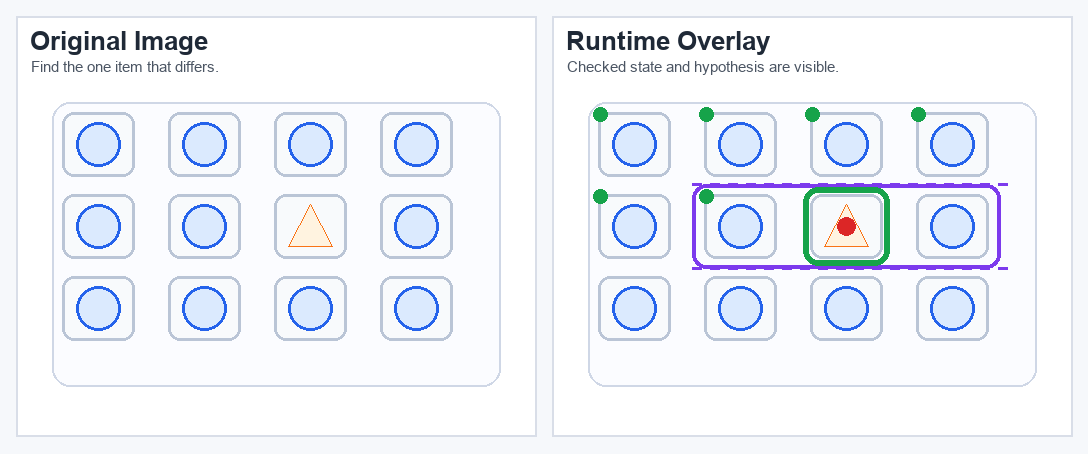}\\[-1pt]
\textcolor{caseMuted}{Checked candidates and the current odd-item hypothesis are rendered as visible search state to avoid repeated comparisons.}
\end{casecard}
\end{minipage}
\caption{\textbf{Dynamic visual skill examples.} Dynamic skills render intermediate state onto the current task image, making progress auditable and reusable across reasoning steps.}
\label{fig:appendix-dynamic-cases}
\end{figure*}

\FloatBarrier
\clearpage

\begin{figure*}[t]
\centering
\scriptsize
\begin{minipage}[t]{0.49\linewidth}
\begin{casecard}{lavenderborder}
\casebadge{caseInterleaved}{INTERLEAVED}\quad\textbf{Pythagorean visual proof}\\[3pt]
\includegraphics[width=\linewidth]{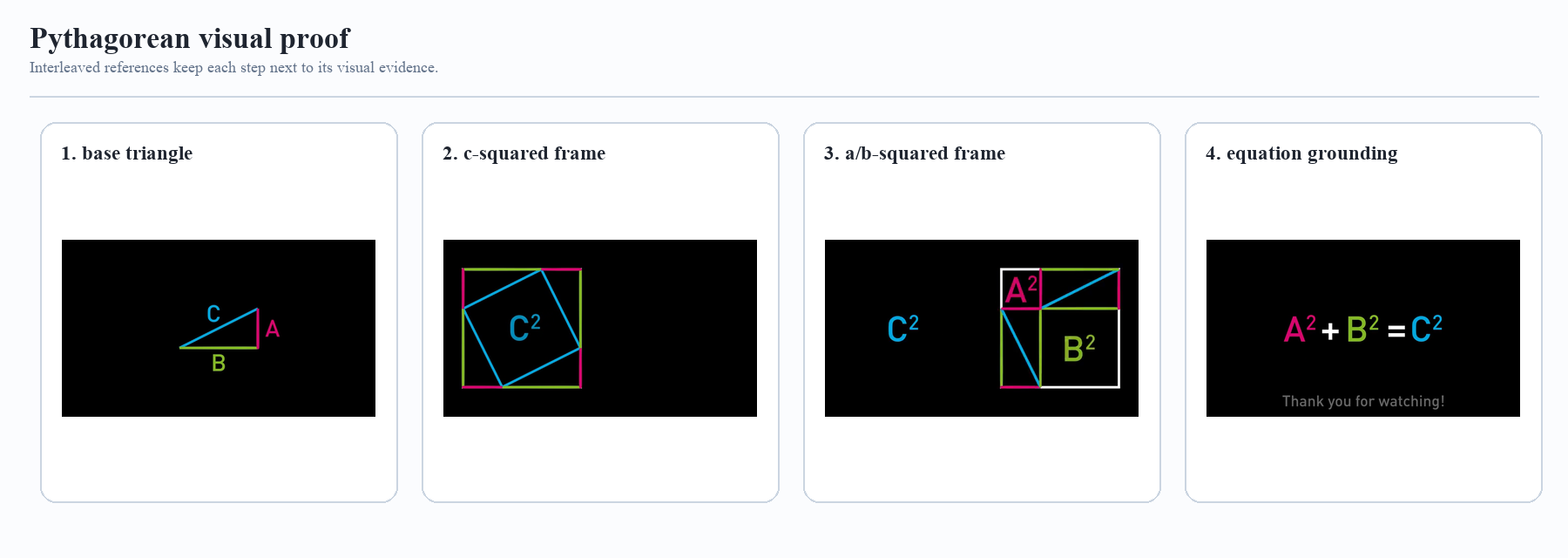}\\[-1pt]
\textcolor{caseMuted}{Sampled video keyframes are bound to the triangle sides, area rearrangements, and final equation. The skill preserves the visual proof as ordered evidence rather than a text-only derivation.}
\end{casecard}
\end{minipage}\hfill
\begin{minipage}[t]{0.49\linewidth}
\begin{casecard}{lavenderborder}
\casebadge{caseInterleaved}{INTERLEAVED}\quad\textbf{VS Code Remote-SSH docs}\\[3pt]
\includegraphics[width=\linewidth]{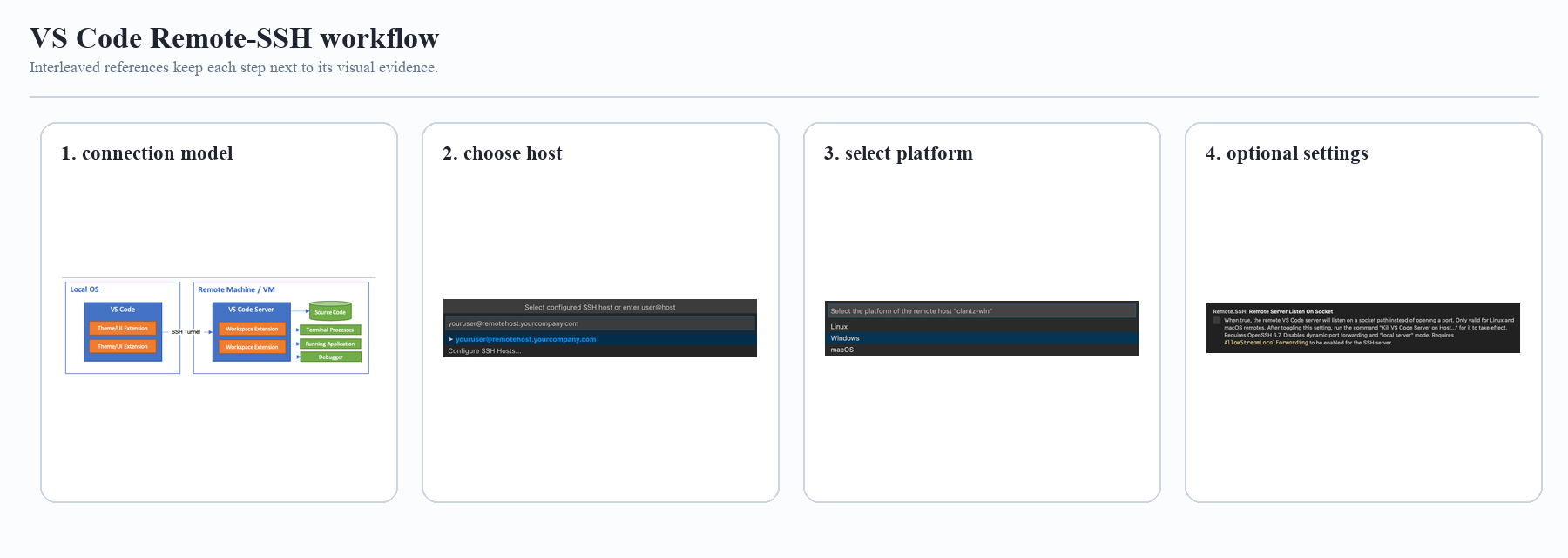}\\[-1pt]
\textcolor{caseMuted}{Documentation screenshots and diagrams are attached to the specific workflow steps they clarify, helping agents and users recognize the relevant UI state.}
\end{casecard}
\end{minipage}
\caption{\textbf{Interleaved visual skill examples.} Interleaved skills preserve ordered text--visual evidence bindings for tutorials, documentation, videos, and document-like sources.}
\label{fig:appendix-interleaved-cases}
\end{figure*}

%%%%%%%%%%%%%%%%%%%%%%%%%%%%%%%%%%%%%%%%%%%%%%%%%%%%%%%%%%%%

\end{document}